\pdfoutput=1

\documentclass[11pt]{article}

\usepackage[]{EMNLP2023}

\usepackage{amsmath,amssymb}
\usepackage{mathtools}
\usepackage{cleveref}
\usepackage{comment}
\usepackage[export]{adjustbox}
\usepackage{relsize}
\usepackage{subfig}
\usepackage{booktabs}
\usepackage{multicol}
\usepackage{multirow}
\usepackage{supertabular}
\usepackage{numprint}
\usepackage{ifthen}
\usepackage{tikz}
\usetikzlibrary{positioning}
\usetikzlibrary{calc}
\usetikzlibrary{arrows.meta}
\usetikzlibrary{shapes}
\usetikzlibrary{decorations}
\usetikzlibrary{fit}
\usepackage{pgfplots}
\pgfplotsset{compat=newest}
\usepgfplotslibrary{colorbrewer}
\usepackage{color}
\usepackage{xcolor}
\definecolor{color1}{HTML}{1b9e77} 
\definecolor{color2}{HTML}{d95f02} 
\definecolor{color3}{HTML}{7570b3} 
\definecolor{color4}{HTML}{e7298a} 
\definecolor{color5}{HTML}{66a61e} 

\definecolor{colorA}{HTML}{00a7d6}
\definecolor{colorB}{HTML}{d9ac36}
\definecolor{colorC}{HTML}{d04f2c}

\usepackage{times}
\usepackage{latexsym}

\usepackage[T1]{fontenc}

\usepackage[utf8]{inputenc}

\usepackage{microtype}

\usepackage{inconsolata}

\title{Why bother with geometry? \\ On the relevance of linear decompositions of Transformer embeddings}

\author{Timothee Mickus \\
  University of Helsinki \\
  \texttt{timothee.mickus@helsinki.fi} \And
  Raúl Vázquez \\
  University of Helsinki \\
  \texttt{raul.vazquez@helsinki.fi} \\}

\begin{document}
\maketitle
\begin{abstract}
A recent body of work has demonstrated that Transformer embeddings can be linearly decomposed into well-defined sums of factors, that can in turn be related to specific network inputs or components.
There is however still a dearth of work studying whether these mathematical reformulations are empirically meaningful.
In the present work, we study representations from machine-translation decoders using two of such embedding decomposition methods. 
Our results indicate that, while decomposition-derived indicators effectively correlate with model performance, variation across different runs suggests a more nuanced take on this question.
The high variability of our measurements indicate that geometry reflects model-specific characteristics more than it does sentence-specific computations, and that similar training conditions do not guarantee similar vector spaces.

\end{abstract}\section{Introduction}

It stands to reason that important research efforts are being devoted to explaining the behavior and understanding the mechanics of Transformer-based NLP models: Most models that are currently discussed within the NLP community are based on this architecture, and they have achieved resounding successes.
One trend of work in particular attempts to characterize Transformer models by means of their geometry \citep[e.g.,]{rogers-etal-2020-primer, ethayarajh-2019-contextual,timkey-van-schijndel-2021-bark}.

However, most studies focus on a single handful of `foundation' models or fine-tuned variants thereof, and explicitly or implicitly assume that the reported results generalize on to other models---yet the effects of random initialization, training data or variation in the definition of objective functions are left unstudied.
Moreover, and perhaps more crucially, there is no guarantee that Transformer-embeddings geometry is indicative of model quality: 
That embeddings are arranged in a certain fashion in hyperspace says little of what downstream performance we should expect.

Taken together, these two assumptions---that results applicable to one model will apply to many, and that geometry can provide explanations---call into question the validity of geometry-based approaches.
We focus on two linear decomposition approaches for Transformer embeddings \citep{mickus-etal-2022-dissect, oh-schuler-2023-token}: works attempting to summarize model computation through their effects on the resulting output embedding spaces. 
By construction, these decompositions reflect topological features of the Transformer architecture. 

Our goal is to verify whether these two assumptions are in fact supported.
One would hope, for instance, that the geometry a model settles on differs along training data but not random initialization. 
Another natural expectation to have is that different uses of the model, such as forcing the production of a given sentence or searching for a plausible generation would yield distinct computations and therefore distinct geometric arrangements.
Lastly, if we intend to explain model performance via embedding geometry, then we should observe consistent differences in geometry whenever we see differences in quality metrics.

To answer whether all three of these expectations are met, we experiment with machine-translation decoder embeddings and study how their geometry evolves over training and multilingualism.
In a nutshell, our experiments suggest a nuanced outlook on the usefulness of linear decompositions. 
Decomposition-derived indicators tend to correlate well with corpus-level model performance, but are less appropriate when it comes to sentence-level performance. 
We also observe that variation in geometry across different runs for a same translation task can exceed what we observe for models trained for different translation tasks.

Our findings question the relevance of geometry-based approaches for Transformer model explainability.
As our measurements display high variability across different model training runs, this work suggests that geometry reflects model-specific characteristics more than it does sentence-specific computations:
Models trained in similar conditions need not yield similar vector spaces.

\section{Related works} \label{sec:sota}

There is a rich literature that connects the objective of static embedding models such as word2vec to characteristics of the resulting vector space.
In particular, \citet{pmlr-v97-allen19a} worked out how the linguistic regularities found by \citet{mikolov-etal-2013-linguistic} result from the exact loss landscape. 

As for contextual embeddings, research has been more commonly limited to empirical observations \citep[e.g.]{ethayarajh-2019-contextual,timkey-van-schijndel-2021-bark}.
Recently, \citet{ferrando-etal-2022-measuring,ferrando-etal-2022-towards,modarressi-etal-2022-globenc,mickus-etal-2022-dissect,oh-schuler-2023-token,yang-etal-2023-local} and others have developed methods to provide mechanistic interpretations of Transformer outputs:
These works rely on linear algebra to derive mathematically exact attributions, where a contextual embedding is decomposed as a sum of interpretable vector terms. 

These approaches build upon two peculiarities of the Transformer architecture.
Perhaps the most famous one---at least, one that has found significant traction more generally across explainable NLP---is that of the scaled dot \emph{attention mechanism}. 
Transformers were presented by \citet{NIPS2017_3f5ee243} as attention-only models. 
Attention mechanisms can be seen as weighted sums over value vectors \citep{kobayashi-etal-2020-attention}, where the attention weights are derived non-linearly.
This observation was first brought forth within the sustained and ongoing discussion about the relevance of attention weights, and whether they are efficient means of explaining Transformer behaviors---a subject hotly debated \citep{jain-wallace-2019-attention,wiegreffe-pinter-2019-attention,pruthi-etal-2020-learning}.
In particular, \citet{serrano-smith-2019-attention,kobayashi-etal-2020-attention} highlight the importance of considering the full embedding space geometry. 

The second characteristic of importance to Transformer decomposition approaches is the systematic use of residual connections throughout a Transformer models, only interrupted by layer normalization operations. 
This fact, often described as a \emph{residual stream} of information,  has been leveraged to interpret the behavior of feed-forward sub-layers \citep{geva-etal-2021-transformer,ferrando-etal-2023-explaining,dar-etal-2023-analyzing} or layer commutativity \citep{9533563}.
Given that on the one hand a layer norm is a linear map, and on the other hand a residual connection simply consists on adding a sub-network's input to its output, 
this entails that most of the computations done in a Transformer are distributive.


\section{Methodology}
\label{sec:methodology}

Our focus here is on sequence-to-sequence encoder-decoder architectures \citep{NIPS2017_3f5ee243}.
Transformer embeddings can be decomposed into a linear combination of nonlinear transformations using properties of the residual connections and attention mechanisms. 
Here, we focus on whether these decompositions do provide meaningful explanations, or whether they merely reflect topological characteristics of the Transformer architecture.\footnote{
    Code at \href{https://github.com/TimotheeMickus/seq2seq-splat}{\tt github.com/TimotheeMickus/seq2seq-splat}.
}

\subsection{Models \& Data}
\label{sec:methodology:models}

Connecting with previous literature \citep[e.g.]{voita-etal-2021-analyzing,ferrando-etal-2022-towards,vazquez-etal-2022-closer}, our focus in this work is on decoder embeddings from Transformers trained on machine translation (MT) objectives, with varying degrees of multilinguality.
NMT systems provide a useful framework to study the validity of explainability methodologies. 
First, significant efforts have been devoted to the creation of MT evaluation metrics that correlate well with human intuitions. 
Empirical investigations of what drives phenomena such as hallucinations also abound.
Lastly, translation as a task has the advantage that is straightforward for humans to relate input to output.

Our models are trained on different subsets of the Tatoeba Challenge corpus \citep{tiedemann-2020-tatoeba}, each of them sampling up to 5M sentences per language pair. 
We train models with sources of different degrees of multilinguality: multilingual-to-English, with 76M sentences; Indo-European-to-English, with 58M sentences; Slavic-to-English, with 33M sentences; and Russian-to-English with 5M sentences. For the bilingual dataset (Ru--En), we train three different model seeds. 
All models are trained using the marian-MT library \citep{junczys-dowmunt-etal-2018-marian} for 72 hours on 4 V100 GPUs. 
We saved checkpoints every 1000 training steps to compare decompositions at different training stages.
Hyperparameters and training details are listed in \cref{adx:hparams}.
We systematically run all of our experiments on the same held out test set of 19,425 Russian and English paired sentences.

\subsection{Decomposition approaches}
\label{sec:methodology:dcps}

\begin{table}[th]
    \centering
    \subfloat[General notations]{\resizebox{0.9\linewidth}{!}{\smaller
    \begin{tabular}{| p{0.125\linewidth} p{0.725\linewidth} |}
        \hline
        $\mathbf{Z}$ & matrix \\
        $\left(\mathbf{Z}\right)_i$ & $i$\textsuperscript{th} row of $\mathbf{Z}$ \\
        $\mathbf{z}$ & (row) vector \\ 
        $k$, $\kappa$, $K$ & scalars \\
        $\mathbf{y}\oplus\mathbf{z}$ & concatenation of vectors $\mathbf{y}$ and $\mathbf{z}$ \\ 
        $\bigoplus\limits_n \mathbf{z}_n$ & $\mathbf{z}_1\oplus\mathbf{z}_2\oplus\dots\oplus\mathbf{z}_n$\\
        $\mathbf{y}\odot\mathbf{z}$ & element-wise multiplication of $\mathbf{y}$ and $\mathbf{z}$ \\ 
        $\bigodot\limits_n \mathbf{z}_n$ & $\mathbf{z}_1\odot\mathbf{z}_2\odot\dots\odot\mathbf{z}_n$\\
        $\vec{1}$ & vector with all components set to 1 \\
        $\vec{0}$ & vector with all components set to 0 \\
        $\mathbf{0}_{m,n}$ & null matrix of shape $m \times n$ \\
        $\mathbf{I}_n$ & identity matrix of shape $n \times n$ \\ \hline
    \end{tabular}
    }}
    
    \subfloat[Transformer-specific notations]{\resizebox{0.9\linewidth}{!}{\smaller
    \begin{tabular}{| p{0.125\linewidth} p{0.725\linewidth} |}
        \hline
        $\Lambda$ & total number of sub-layers \\
        $\lambda$ & sub-layer index \\
        $L$ & total number of layers, i.e., $\Lambda/3$ \\
        $l$ & layer index \\
        $d$ & dimension of representations \\
        $H$ & number of heads \\
        $\mathbf{W}^{(\mathrm{m})}_\lambda$ & sub-module $\mathrm{m}$ in sub-layer~$\lambda$ weight matrix  \\ 
        $\mathbf{b}^{(\mathrm{m})}_\lambda$ & bias for sub-module $\mathrm{m}$ in sub-layer $\lambda$ \\ 
        $\mathbf{g}^{(\mathrm{ln})}_\lambda$ & layer-norm gain parameter in sub-layer $\lambda$ \\ 
        $\mathbf{E}_\lambda$ & output of sub-layer $\lambda$ (all embeddings) \\
        $\mathbf{e}_{\lambda,t}$ & output of sub-layer $\lambda$ at position $t$ \\
        $\mathbf{\dot{e}}_{\lambda,t}$ & output of sub-layer $\lambda$ at position $t$ before layer-norm and residual connection \\
        $\mathbf{\ddot{e}}_{\lambda,t}$ & output of sub-layer $\lambda$ at position $t$ before layer-norm \\
        $\mathbf{X}_\lambda$ & target-side input to sub-layer $\lambda$, $\mathbf{E}_{\lambda -1}$\\
        $\mathbf{x}_{\lambda,t}$ & $t$\textsuperscript{th} target-side input of sub-layer $\lambda$, $\mathbf{e}_{\lambda -1}$\\
        $\mathbf{X}_{\mathrm{enc}}$ & memory bank, i.e., output of the encoder\\
        $\mathbf{A}_{\lambda,h}$ & attention weight matrix for $h$\textsuperscript{th} head of the multi-head attention at sublayer $\lambda$\\ 
        $a_{\lambda htt'}$ & Attention weight for head $h$, sub-layer $\lambda$ query $t$, value $t'$\\
        $\phi$ & non-linear activation function\\ 
        $m_{\lambda t}$ & mean from the layer-norm of sub-layer $\lambda$\\
        $s_{\lambda t}$ & standard deviation from $\lambda$\textsuperscript{th} layer-norm \\
        \hline
    \end{tabular}
    }}
    \caption{Notations}
    \label{tab:notations}
\end{table}

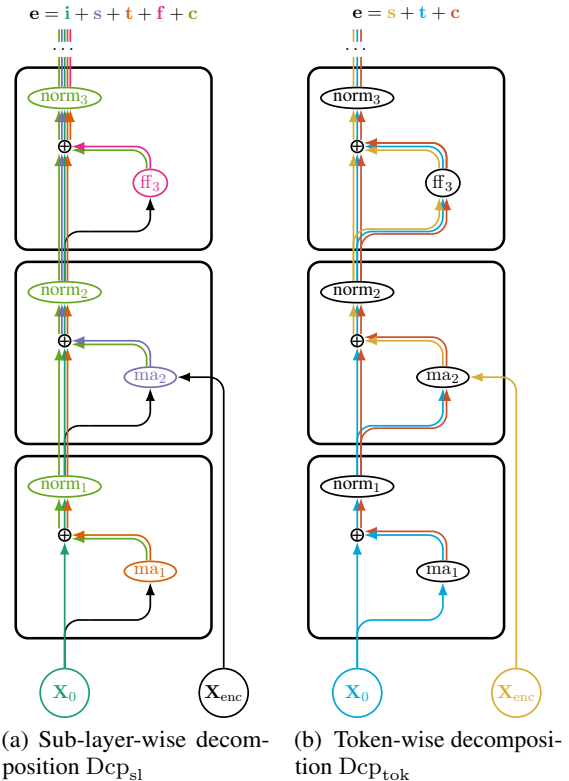
\begin{figure}[th]
    \centering
    \subfloat[\label{fig:overview:sl}Sub-layer-wise decomposition $\mathrm{Dcp}_{\mathrm{sl}}$]{
        \resizebox{0.425\columnwidth}{!}{
            \begin{tikzpicture}[
>=LaTeX,
cell/.style={
    rectangle,
    rounded corners=2.5mm,
    draw,
    line width = 1.5pt,
    },
operator/.style={
    circle,
    draw,
    inner sep=-0.5pt,
    line width = 1pt,
    minimum height =.2cm,
    },
function/.style={
    ellipse,
    draw,
    line width = 1pt,
    inner sep=1pt
    },
ct/.style={
    circle,
    draw,
    line width = 1pt,
    minimum width=1cm,
    inner sep=.5pt,
    },
gt/.style={
    rectangle,
    draw,
    minimum width=4mm,
    minimum height=3mm,
    inner sep=1pt
    },
mylabel/.style={
    font=\scriptsize\sffamily
    },
ArrowC1/.style={
    rounded corners=.3cm,
    line width=1pt,
    },
ArrowC2/.style={
    rounded corners=.75cm,
    line width=1pt,
    },
]

\def\layerscale{4}
\def\layeroffset{0}
\foreach \layer [count=\li] in {{$\mathrm{ma}_1$},{$\mathrm{ma}_2$},{$\mathrm{ff}_3$}}
{
  \node [cell, minimum height =3.75cm, minimum width=4cm] at (-1.5,\li * \layerscale + \layeroffset){} ;
  \ifthenelse{\li = 1}{
    \node [function, color2] (f\li) at (-0.75,-0.5 + \li * \layerscale + \layeroffset) {\layer};
  }{
    \ifthenelse{\li = 2}{
      \node [function, color3] (f\li) at (-0.75,-0.5 + \li * \layerscale + \layeroffset) {\layer};
    }{
      \node [function, color4] (f\li) at (-0.75,-0.5 + \li * \layerscale + \layeroffset) {\layer};
    }
  }
  \node [] (resid\li) at (-2.5,0.+ \li * \layerscale + \layeroffset) {};
  \node [operator, ] (sum\li) at (-2.5,0.25 + \li * \layerscale + \layeroffset) {+};
  \node [] (bend\li) at (-2, -1.5 + \li * \layerscale + \layeroffset) {};
  \node [function, color5] (n\li) at (-2.5,1.25 + \li * \layerscale + \layeroffset) {$\text{norm}_\li$};

  \draw [->, ArrowC1] (bend\li.center) -| (f\li);
  \draw [->, ArrowC1, color5] ($(f\li.north) + (-0.075, 0)$) |- ($(sum\li.east) + (0, -0.075)$);
  \ifthenelse{\li = 1}{
    \draw [->, ArrowC1, color2] (f\li) |- (sum\li);
  }{
    \ifthenelse{\li = 2}{
      \draw [->, ArrowC1, color3] (f\li) |- (sum\li);
    }{
      \draw [->, ArrowC1, color4] (f\li) |- (sum\li);
    }
  }
  \draw [->, ArrowC2, color1] (sum\li) -- (n\li);
  \draw [->, ArrowC2, color2] ([xshift=0.15em]sum\li.north) -- ([xshift=0.15em]n\li.south);
  \draw [->, ArrowC2, color5] ([xshift=-0.3em]sum\li.north) -- ([xshift=-0.3em]n\li.south);
    
  \ifthenelse{\li = 1}{}{
    \draw [->, ArrowC2, color3] ([xshift=-0.15em]sum\li.north) -- ([xshift=-0.15em]n\li.south);
  }
  \ifthenelse{\li = 3}{
    \draw [->, ArrowC2, color2] ([xshift=0.3em]sum\li.north) -- ([xshift=0.3em]n\li.south);
  }{}
  
  \draw [->, ArrowC2, color1] (resid\li.center) -- (sum\li);
  \ifthenelse{\li = 2 \OR \li = 3}{
    \draw [->, ArrowC2, color2] ([xshift=0.15em]resid\li.center) -- ([xshift=0.15em]sum\li.south);
    \draw [->, ArrowC2, color5] ([xshift=-0.3em]resid\li.center) -- ([xshift=-0.3em]sum\li.south);
  }{};
  \ifthenelse{\li = 3}{\draw [->, ArrowC2, color3] ([xshift=-0.15em]resid\li.center) -- ([xshift=-0.15em]sum\li.south);}{};

}
\foreach \li [count=\prevli from 1] in {2, 3}
{
  \draw [ArrowC1] (n\prevli) |- (bend\li.center);
  \draw [ArrowC1, color1] (n\prevli.north) -- (resid\li.center);
  \draw [ArrowC1, color2] ([xshift=0.15em]n\prevli.north) -- ([xshift=0.15em]resid\li.center);
  \draw [ArrowC1, color5] ([xshift=-0.3em]n\prevli.north) -- ([xshift=-0.3em]resid\li.center);
  \ifthenelse{\prevli = 2} {
    \draw [ArrowC1, color3] ([xshift=-0.15em]n\prevli.north) -- ([xshift=-0.15em]resid\li.center);
  }{};
}

\node [] (all) at (-2.5,6.25 + 2 * \layerscale) {$\dots$};
\node[] at (-1.5,7 + 2 * \layerscale) {$\mathbf{e} = \mathbf{\color{color1}i} + \mathbf{\color{color3}s} + \mathbf{\color{color2}t} + \mathbf{\color{color4}f} + \mathbf{\color{color5}c}$};
\node [] (o) at (-2.5,6.8 + 2 * \layerscale) {};

\node[ct, color1] (x) at (-2.5,1.0) {$\mathbf{X}_0$};
\node[ct] (xenc) at (.75,1.0) {$\mathbf{X}_\mathrm{enc}$};
\draw [ArrowC1] (x) |- (bend1.center);
\draw [ArrowC1, color1] (x)-- (resid1.center);
\draw [->, ArrowC1] (xenc) |- (f2.east);

\draw [ArrowC2, color1] (n3) -- (all) -- (o);
\draw [ArrowC2, color2] ([xshift=0.15em]n3.north) -- ([xshift=0.15em]all.south);
\draw [ArrowC2, color2] ([xshift=0.15em]all.north) -- ([xshift=0.15em]o.south);
\draw [ArrowC2, color3] ([xshift=-0.15em]n3.north) -- ([xshift=-0.15em]all.south);
\draw [ArrowC2, color3] ([xshift=-0.15em]all.north) -- ([xshift=-0.15em]o.south);
\draw [ArrowC2, color4] ([xshift=0.3em]n3.north) -- ([xshift=0.3em]all.south);
\draw [ArrowC2, color4] ([xshift=0.3em]all.north) -- ([xshift=0.3em]o.south);
\draw [ArrowC2, color5] ([xshift=-0.3em]n3.north) -- ([xshift=-0.3em]all.south);
\draw [ArrowC2, color5] ([xshift=-0.3em]all.north) -- ([xshift=-0.3em]o.south);
\end{tikzpicture}
        }
    }
    \quad
    \subfloat[\label{fig:overview:tok} Token-wise decomposition $\mathrm{Dcp}_{\mathrm{tok}}$]{
        \resizebox{0.425\columnwidth}{!}{
            \begin{tikzpicture}[
>=LaTeX,
cell/.style={
    rectangle,
    rounded corners=2.5mm,
    draw,
    line width = 1.5pt,
    },
operator/.style={
    circle,
    draw,
    inner sep=-0.5pt,
    line width = 1pt,
    minimum height =.2cm,
    },
function/.style={
    ellipse,
    draw,
    line width = 1pt,
    inner sep=1pt
    },
ct/.style={
    circle,
    draw,
    line width = 1pt,
    minimum width=1cm,
    inner sep=.5pt,
    },
gt/.style={
    rectangle,
    draw,
    minimum width=4mm,
    minimum height=3mm,
    inner sep=1pt
    },
mylabel/.style={
    font=\scriptsize\sffamily
    },
ArrowC1/.style={
    rounded corners=.3cm,
    line width=1pt,
    },
ArrowC2/.style={
    rounded corners=.75cm,
    line width=1pt,
    },
]

\def\layerscale{4}
\def\layeroffset{0}
\foreach \layer [count=\li] in {{$\mathrm{ma}_1$},{$\mathrm{ma}_2$},{$\mathrm{ff}_3$}}
{
  \node [cell, minimum height =3.75cm, minimum width=4cm] at (-1.5,\li * \layerscale + \layeroffset){} ;
  \node [function] (f\li) at (-0.75,-0.5 + \li * \layerscale + \layeroffset) {\layer};
  \node [] (resid\li) at (-2.5,0.+ \li * \layerscale + \layeroffset) {};
  \node [operator, ] (sum\li) at (-2.5,0.25 + \li * \layerscale + \layeroffset) {+};
  \node [] (bend\li) at (-2, -1.5 + \li * \layerscale + \layeroffset) {};
  \node [function] (n\li) at (-2.5,1.25 + \li * \layerscale + \layeroffset) {$\text{norm}_\li$};

    \node[] (bend2\li) at ($(bend\li) + (.05, .05)$)  {};
    \node[] (bend3\li) at ($(bend\li) + (-.05, -.05)$)  {};

  \draw [->, ArrowC1, colorA] (bend\li.center) -| (f\li);
  \draw [->, ArrowC2, colorA] (sum\li) -- (n\li);
  \draw [->, ArrowC2, colorC] ([xshift=0.2em]sum\li.north) -- ([xshift=0.2em]n\li.south);
  \draw [->, ArrowC1, colorC] ($(f\li.north) + (0.075, 0)$) |- ($(sum\li.east) + (0, 0.075)$);
  
  \ifthenelse{\li = 1}{
    \draw [->, ArrowC1, colorA] (f\li) |- (sum\li);
    \draw [->, ArrowC2, colorA] (resid\li.center) -- (sum\li.south);
  }{
    \draw [->, ArrowC1, colorC] (bend3\li.center) -| ($(f\li.south) + (0.075, 0)$);
  }

  \ifthenelse{\li = 2}{
    \draw [->, ArrowC1, colorB] (f\li) |- (sum\li);
    \draw [->, ArrowC2, colorB] ([xshift=-0.2em]sum\li.north) -- ([xshift=-0.2em]n\li.south);
    \draw [->, ArrowC2, colorA] (resid\li.center) -- (sum\li.south);
    \draw [->, ArrowC2, colorC] ([xshift=0.2em]resid\li.center) -- ([xshift=0.2em]sum\li.south);
  }{}
  
  \ifthenelse{\li = 3}{
    \draw [->, ArrowC1, colorA] (f\li) |- (sum\li);
    \draw [->, ArrowC1, colorB] ($(f\li.north) + (-0.075, 0)$) |- ($(sum\li.east) + (0, -0.075)$);
    \draw [->, ArrowC1, colorC] ($(f\li.north) + (0.075, 0)$) |- ($(sum\li.east) + (0, 0.075)$);
    \draw [->, ArrowC2, colorB] ([xshift=-0.2em]sum\li.north) -- ([xshift=-0.2em]n\li.south);
    \draw [->, ArrowC2, colorA] (resid\li.center) -- (sum\li.south);
    \draw [->, ArrowC2, colorC] ([xshift=0.2em]resid\li.center) -- ([xshift=0.2em]sum\li.south);
    \draw [->, ArrowC2, colorB] ([xshift=-0.2em]resid\li.center) -- ([xshift=-0.2em]sum\li.south);
  
  }{}

}
\foreach \li [count=\prevli from 1] in {2, 3}
{
  \draw [ArrowC1, colorA] (n\prevli) |- (bend\li.center);
  \draw [ArrowC1, colorA] (n\prevli.north) -- (resid\li.center);
  \draw [ArrowC1, colorC] ([xshift=0.2em]n\prevli.north) -- ([xshift=0.2em]resid\li.center);
  \draw [ArrowC1, colorC] ([xshift=0.2em]n\prevli.north) |- (bend3\li.center);
  \ifthenelse{\prevli = 2} {
    \draw [ArrowC1, colorB] ([xshift=-0.2em]n\prevli.north) -- ([xshift=-0.2em]resid\li.center);
    \draw [ArrowC1, colorB] ([xshift=-0.2em]n\prevli.north) |- (bend2\li.center);
    \draw [->, ArrowC1, colorB] (bend2\li.center) -| ($(f\li.south) + (-0.075, 0)$);
  }{};
}

\node [] (all) at (-2.5,6.25 + 2 * \layerscale) {$\dots$};
\node[] at (-1.5,7 + 2 * \layerscale) {$\mathbf{e} = \mathbf{\color{colorB}s} + \mathbf{\color{colorA}t} + \mathbf{\color{colorC}c}$};
\node [] (o) at (-2.5,6.8 + 2 * \layerscale) {};

\node[ct, colorA] (x) at (-2.5,1.0) {$\mathbf{X}_0$};
\node[ct, colorB] (xenc) at (.75,1.0) {$\mathbf{X}_\mathrm{enc}$};
\draw [ArrowC1, colorA] (x) |- (bend1.center);
\draw [ArrowC1, colorA] (x)-- (resid1.center);
\draw [->, ArrowC1, colorB] (xenc) |- (f2.east);

\draw [ArrowC2, colorA] (n3) -- (all) -- (o);
\draw [ArrowC2, colorB] ([xshift=-0.2em]n3.north) -- ([xshift=-0.2em]all.south);
\draw [ArrowC2, colorB] ([xshift=-0.2em]all.north) -- ([xshift=-0.2em]o.south);
\draw [ArrowC2, colorC] ([xshift=0.2em]n3.north) -- ([xshift=0.2em]all.south);
\draw [ArrowC2, colorC] ([xshift=0.2em]all.north) -- ([xshift=0.2em]o.south);
\end{tikzpicture}
        }
    }
    \caption{Overview of decomposition methods, focusing on the first three sublayers of the decoder. Colors indicate what a decomposition term is imputed to.}
    \label{fig:overview}
\end{figure}

We consider two approaches: a sub-layer-wise decomposition, and a token-wise decomposition.
They are inspired by \citet{mickus-etal-2022-dissect} and \citet{oh-schuler-2023-token} and illustrated in \cref{fig:overview}. 
In \cref{tab:notations}, we list the notations used throughout this work.
See \cref{adx:dcp-supinfo} for a primer on the Transformer architecture.

\paragraph{Sub-layer-wise decomposition.}
The first approach is a sub-layer-level decomposition in five terms. That is, we decompose embedding $\mathbf{e}$ into a linear combination of functions that refer to the target-side input ($\mathbf{i}$), the source attention ($\mathbf{s}$), the target attention ($\mathbf{t}$), the feed-forwards ($\mathbf{f}$), or the models' biases ($\mathbf{c}$).
We note it as $\mathrm{Dcp}_{\mathrm{sl}}$.
As can be seen in \cref{fig:overview:sl}, it essentially entails that we break down embeddings depending on where in the network a specific term comes from.
Hence, for token position $t$: 
\begin{equation}
 \mathbf{e}_t = \mathbf{i}_t + \mathbf{s}_t + \mathbf{t}_t + \mathbf{f}_t + \mathbf{c}_t    
\end{equation}

where
\begin{align}
    \mathbf{i}_t &= f^{(\mathrm{ln})}_1 \left(\mathbf{x}_{0,t} \right) \\
    \mathbf{t}_t &= \sum_{l=0}^{\Lambda/3 - 1} f^{(\mathrm{ln})}_{3l + 1}\left( \left(f_{3l + 1}^{(\mathrm{ma})}\left(\mathbf{X}_{3l + 1}\right)\right)_t \right) \\
    \mathbf{s}_t &= \sum_{l=0}^{\Lambda/3 - 1} f^{(\mathrm{ln})}_{3l + 2}\left( \left(f_{3l + 2}^{(\mathrm{ma})}\left(\mathbf{X}_{\mathrm{enc}}\right)\right)_t \right) \\
    \mathbf{f}_t &= \sum_{l=0}^{\Lambda/3 - 1} f^{(\mathrm{ln})}_{3l + 3}\left( f_{3l + 3}^{(\mathrm{ff})}\left(\mathbf{x}_{3l + 3,t}\right) \right) 
\end{align}
\vspace{-.5cm}
\begin{equation}%
\resizebox{0.95\linewidth}{!}{%
    $\begin{aligned}\mathbf{c}_t &= \mathbf{b}^{(\mathrm{ln})}_\Lambda + f_1^{(\mathrm{ln})}(-m_1 \vec{1}) + \sum_{\lambda=2}^{\Lambda} f_\lambda^{(\mathrm{ln})}\left(\mathbf{b}^{(\mathrm{ln})}_{\lambda-1} -m_\lambda \vec{1} \right) \\ 
                 & \quad + \sum_{l=0}^{\Lambda / 3 - 1} f_{\lambda+1}^{(\mathrm{ln})}\left( \mathbf{b}_{\lambda+1}^{(\mathrm{ma,O})} +  \sum\limits_{h=1}^H  \mathbf{H}_{\lambda+1, h} \mathbf{b}_{\lambda+1}^{(\mathrm{ma,V})}\right) \\ 
                 & \quad + \sum_{l=0}^{\Lambda / 3 - 1} f_{\lambda+2}^{(\mathrm{ln})}\left( \mathbf{b}_{\lambda+2}^{(\mathrm{ma,O})} +  \sum\limits_{h=1}^H  \mathbf{H}_{\lambda+2, h} \mathbf{b}_{\lambda+2}^{(\mathrm{ma,V})} \right) \\ 
                 & \quad + \sum_{l=0}^{\Lambda / 3 - 1} f_{\lambda+3}^{(\mathrm{ln})}\left( \mathbf{b}_{\lambda+3}^{(\mathrm{ff}, \mathrm{out})} \right) 
    \end{aligned}$
}%
\end{equation}
The cumulative effects of the layer-norms after sub-layer $\lambda$, $f^{(\mathrm{ln})}_\lambda(\mathbf{x})$, the unbiased outputs of a feed-forward layer, $ f_\lambda^{(\mathrm{ff})}\left(\mathbf{x}\right)$, and of a multi-head attention layer, $f_\lambda^{(\mathrm{ma})}\left(\mathbf{X}\right)$,  are defined as follows:
\begin{align*}
    f^{(\mathrm{ln})}_\lambda(\mathbf{x}) &= \frac{1}{\prod \limits_{\lambda'=\lambda}^\Lambda s_{\lambda'}} \bigodot \limits_{\lambda'=\lambda}^\Lambda \mathbf{g}_{\lambda'} \odot \mathbf{x} \\
    f_\lambda^{(\mathrm{ff})}\left(\mathbf{x}\right) &=
        \mathbf{W}^{(\mathrm{ff}, \mathrm{out})}_\lambda \phi\left(\mathbf{W}^{(\mathrm{ff}, \mathrm{in})}_\lambda \mathbf{x} + \mathbf{b}^{(\mathrm{ff}, \mathrm{in})}_\lambda\right)\\
    f_\lambda^{(\mathrm{ma})}\left(\mathbf{X}\right) &= \mathbf{W}^{(\mathrm{ma}, \mathrm{O})}_\lambda \left( \bigoplus\limits_{h=1}^H \mathbf{A}_{\lambda,h} \mathbf{W}^{(\mathrm{ma}, \mathrm{V})}_{\lambda,h} \mathbf{X}\right)
\end{align*}
For convenience we also define the linear map associated with going from a given head $h$ to the output of sub-layer $\lambda$:
\begin{align*}
    \mathbf{H}_{\lambda, h} &= \mathbf{W}^{(\mathrm{ma,O})}_\lambda \mathbf{S}_h \\ 
    \mathbf{S}_h^{} &= \begin{bmatrix}
        \mathbf{0}_{\frac{d}{H},\frac{d(h-1)}{H}} & \mathbf{I}_{\frac{d}{H}} & \mathbf{0}_{\frac{d}{H},\frac{d(H-h)}{H}} 
    \end{bmatrix}    
\end{align*}

\paragraph{Token-wise decomposition.}
A major issue that stands in the way of linear decomposition approaches is the use of a non-linear activation function in feed-forward sub-layers.
This has prompted different approaches: side-stepping the problem altogether and leaving this component unanalyzed \citep{mickus-etal-2022-dissect,ferrando-etal-2022-measuring,modarressi-etal-2022-globenc}; relying on local linear approximations of the activation function \citep{oh-schuler-2023-token}; or limiting the scope of inquiry to activation functions with the desired mathematical properties  \citep{yang-etal-2023-local}.

The second decomposition we study, which we note $\mathrm{Dcp}_{\mathrm{tok}}$, uses the locally linear approximation of \citet{oh-schuler-2023-token} to distribute the feed-forward sub-layer outputs to the input decomposition. 
We then group all inputs into three terms $\mathbf{s}, \mathbf{t}, \mathbf{c}$, depending on whether a vector term ultimately comes from the encoder, from the target input ($\mathbf{t}$) or model biases ($\mathbf{c}$), as shown in \cref{fig:overview:tok}.
Unlike $\mathrm{Dcp}_\mathrm{sl}$, this entails grouping terms based on what they originally were. 
More formally, we define it as:
\begin{equation}
    \mathbf{e}_{\lambda,t} = \mathbf{s}_{\lambda,t} + \mathbf{t}_{\lambda,t} + \mathbf{c}_{\lambda,t} 
\end{equation}
and compute these operands by recurrence.

If we start by setting aside layer normalization and residual connection for simplicity, we can get a first approximation of what should be attributed to the source-side input at a given sub-layer, given prior computations:
\begin{equation}
    \mathbf{\dot{s}}_{\lambda, t} = \begin{cases}
                        \sum\limits_{t'=1}^t a_{\lambda h t t'} \mathbf{H}_{\lambda, h} \mathbf{W}^{(\mathrm{ma,V})}_{\lambda,h}  \mathbf{s}_{\lambda-1, t'}  \\[-0.1cm] \hspace{2.125cm} { \mathrm{if} ~ \lambda \equiv 1 \mod{3} } \\[0.25cm]
                        \sum\limits_n \left(f_\lambda^{(\mathrm{ma})}\left(\mathbf{X}_\mathrm{enc}\right)\right)_n  \\[-0.1cm] \hspace{2.125cm} { \mathrm{if} ~ \lambda \equiv 2 \mod{3}} \\[0.25cm]
                        \mathbf{F}_{\lambda, t} \mathbf{s}_{\lambda-1, t}  \hfill { \mathrm{if} ~ \lambda \equiv 0 \mod{3} } 
                        \end{cases} 
\end{equation}
given the local linear approximation of the feed-forward, $\mathbf{F}_{\lambda, t} = \mathbf{W}^{(\mathrm{ff}, \mathrm{out})}_{\lambda} \mathbf{L}_{\lambda,\mathbf{e}_{\lambda, t}} \mathbf{W}^{(\mathrm{ff}, \mathrm{in})}_{\lambda}$.
The local linear approximation itself $\mathbf{L}_{\lambda,\mathbf{x}} $ of the activation function $\phi$ for sub-layer $\lambda$ is defined as:
\begin{equation*}
    \mathbf{L}_{\lambda,\mathbf{x}} = \mathbf{I}_d \odot \phi'\left(\mathbf{W}^{(\mathrm{ff}, \mathrm{in})}_{\lambda}\mathbf{x} + \mathbf{b}^{(\mathrm{ff}, \mathrm{in})}_{\lambda}\right)
\end{equation*}
We can also remark that in the initial stages, the source-side input is not used, meaning that:
\begin{equation*}
    \mathbf{s}_{0, t} = \vec{0} 
\end{equation*}
    
With an analogous line of thought, we can characterize what in a given sub-layer hidden representation is owed to the target-side input as:
\begin{align}
    \mathbf{\dot{t}}_{\lambda, t} &= \begin{cases}
                        \sum\limits_{t'=1}^t a_{\lambda h t t'} \mathbf{H}_{\lambda, h} \mathbf{W}^{(\mathrm{ma,V})}_{\lambda,h} \mathbf{t}_{\lambda-1, t'}  \\[-0.1cm] \hspace{2.125cm}  { \mathrm{if} ~ \lambda \equiv 1 \mod{3}} \\[0.25cm]
                        \vec{0} \hfill { \mathrm{if} ~ \lambda \equiv 2 \mod{3}} \\[0.25cm]
                        \mathbf{F}_{\lambda, t} \mathbf{t}_{\lambda-1, t} \hfill { \mathrm{if} ~ \lambda \equiv 0 \mod{3} }
                        \end{cases}  \\
    \mathbf{t}_{0, t} &= \mathbf{x}_{0,t} \nonumber
\end{align}

And similarly, we can keep track of all biases and offsets thus far ignored:
\begin{align}
        \mathbf{\dot{c}}_{\lambda, t} &= \begin{cases}
                        \mathbf{b}_{\lambda}^{(\mathrm{ma,O})} +  \sum\limits_{h=1}^H  \mathbf{H}_{\lambda, h} \mathbf{b}_{\lambda,h}^{(\mathrm{ma,V})} \\ ~ + \sum\limits_{t'=1}^t a_{\lambda h t t'} \mathbf{H}_{\lambda, h}  \mathbf{W}^{(\mathrm{ma,V})}_{\lambda,h} \mathbf{c}_{\lambda-1, t'}    \\[-0.1cm] \hspace{2.125cm}  { \mathrm{if} ~ \lambda \equiv 1 \mod{3}} \\[0.25cm]
                        \mathbf{b}_{\lambda}^{(\mathrm{ma,O})} +  \sum\limits_{h=1}^H \mathbf{H}_{\lambda, h} \mathbf{b}_{\lambda,h}^{(\mathrm{ma,V})}   \\[-0.1cm] \hspace{2.125cm} { \mathrm{if} ~ \lambda \equiv 2 \mod{3}} \\[0.25cm]
                          \mathbf{b}_\lambda^{(\mathrm{ff}, \mathrm{out})} + \mathbf{W}^{(\mathrm{ff}, \mathrm{out})}_{\lambda} \mathbf{l}_{\lambda,\mathbf{e}_{\lambda, t}} \\ ~ +  \mathbf{W}^{(\mathrm{ff}, \mathrm{out})}_{\lambda} \mathbf{L}_{\lambda,\mathbf{e}_{\lambda, t}} \mathbf{b}_\lambda^{(\mathrm{ff}, \mathrm{in})} \\ ~ +  \mathbf{F}_{\lambda, t} \mathbf{c}_{\lambda-1, t}   \\[-0.1cm] \hspace{2.125cm} { \mathrm{if} ~ \lambda \equiv 0 \mod{3}} 
                        \end{cases}  \\
    \mathbf{c}_{0, t} &= \vec{0} \nonumber
\end{align}
where the intercept of the local linear approximation of the feed-forward activation is defined as:
\begin{align*}
     \mathbf{l}_{\lambda,\mathbf{e}_{\lambda, t}} &= \phi\left(\mathbf{\hat{e}}_{\lambda,t}\right) - \mathbf{L}_{\lambda,\mathbf{e}_{\lambda, t}} \mathbf{\hat{e}}_{\lambda,t} \\
     \mathbf{\hat{e}}_{\lambda,t} &= \mathbf{W}_\lambda^{(\mathrm{ff,in})} \mathbf{e}_{\lambda, t} + \mathbf{b}_\lambda^{(\mathrm{ff,in})}
\end{align*}

Finally, we need to account for residual connections and layer normalisation so as to obtain the exact decomposition for the next layer:
\footnote{
    As our interest lies in disentangling source and target-side contributions, the decomposition above does not properly attribute weights to individual tokens, i.e., all inputs are not disentangled.
    Also remark that the local linear approximation $\mathbf{L}_{\lambda,\mathbf{x}}$ is defined with respect to the hidden state $\mathbf{e}_{\lambda, t}$: 
        As such, the computations it describes are specific to a particular contextualized embedding, which obfuscates token-level attribution.
}
\begin{align}
        \mathbf{s}_{\lambda,t} &= \frac{1}{s_{\lambda, t}} \mathbf{g}_\lambda \odot \left(\mathbf{\dot{s}}_{\lambda,t} + \mathbf{s}_{\lambda - 1,t} \right) \\
        \mathbf{t}_{\lambda,t} &= \frac{1}{s_{\lambda, t}} \mathbf{g}_\lambda \odot \left(\mathbf{\dot{t}}_{\lambda,t} + \mathbf{t}_{\lambda - 1,t} \right) \\ 
        \mathbf{c}_{\lambda,t} &= \frac{1}{s_{\lambda, t}} \mathbf{g}_\lambda \odot \left(\mathbf{\dot{c}}_{\lambda,t} + \mathbf{c}_{\lambda - 1,t} - m_{\lambda,t}\vec{1} \right) \nonumber \\ & \qquad + \mathbf{b}^{(\mathrm{ln})}_\lambda
\end{align}

\subsection{Scalar indicators}
\label{sec:methodology:metrics}

Linear decomposition approaches, by design, yield sums of high-dimensional vectors. To reduce these vectors to comprehendable scalars, we consider two scalar-valued indicator metrics: one that evaluates the relative magnitude magnitude of a term in a linear decomposition with respect to the total embedding; and a cosine-based one as an indicator of co-directionality. 
We choose these indicators due to their simplicity and interpretability.

We define the \emph{norm ratio} as the ratio of $l_2$ norms so as to capture a sense of scale, and the cosine similarity as:
\begin{align} 
    \mathrm{nr}\left(\mathbf{z}, \mathbf{e}\right) &= \frac{\lVert \mathbf{z} \rVert_2^{}}{\lVert \mathbf{e} \rVert_2^{}}  \label{eq:nr} \\
    \cos\left(\mathbf{z}, \mathbf{e}\right) &= \frac{\mathbf{z} \cdot \mathbf{e}}{\lVert \mathbf{z} \rVert_2^{} \lVert \mathbf{e} \rVert_2^{}}  \label{eq:cos}
\end{align}

Intuitively, if a term $\mathbf{z}$ in some decomposition $\mathrm{Dcp}$ of a contextual embedding $\mathbf{e}$ has a small norm, then we should expect this term $\mathbf{z}$ to be unimportant as it effectively contributes little to the total embedding $\mathbf{e}$, resulting in a small norm ratio. 
On the other hand, when a term $\mathbf{z}$ has a large norm, this measure assigns importance to it, regardless of its orientation  with respect to the total embedding $\mathbf{e}$. 
This is instead captured through cosine similarity: 
Co-directionality indicates whether a term $\mathbf{z}$ is pointing in the same direction as the total embedding $\mathbf{e}$ (when $\cos\left(\mathbf{z}, \mathbf{e}\right) = 1$) or in the opposite direction (when $\cos\left(\mathbf{z}, \mathbf{e}\right) = -1$). 
Cosine similarity has long been used in IR and embedding research \citep{singhal2001}.
Taken together, the two indicators  provide a more complete picture, allowing interpretations while retaining simplicity.\footnote{
    In preliminary experiments, we also experimented with Euclidean distance as well as the scalar product importance metric $\mu$ of \citet{mickus-etal-2022-dissect}, eq.~6. 
    We do not include them in the present article for simplicity.
    Also remark that for all decomposition term $\mathbf{z}$ of a given embedding $\mathbf{e}$, we have  $\cos(\mathbf{z}, \mathbf{e}) \mathrm{nr} (\mathbf{z}, \mathbf{e}) = \mu(\mathbf{z}, \mathbf{e})$
}

\section{What is geometry indicative of?}
Given our experimental protocol described in \cref{sec:methodology}, we now explore what is encoded in linear decomposition terms.

\paragraph{Do the decoding algorithms affect the geometry of embeddings?} 
The first element we consider is whether \emph{forced inference}, where we feed the gold target to the model, and a \emph{beam-search} decoding produce different embeddings, as far as a linear decomposition would capture it.
We consider these two decoding algorithms, as they are commonly used in MT studies; moreover we strongly expect that they should entail different behaviors and information flows through the network:
Forced decoding uses a gold reference translation in addition to the source sentence, while beam search doesn't receive this input but instead is a mode searching heuristic. 
It is sensible to expect that the decomposition of embeddings from both decoding algorithms differ. 

In particular, it makes sense to consider how forced inference and beam-search decoding evolve across training.
Models are trained to optimize the likelihood on iid. data:
As such, differences between these two decoding algorithms---if any are to be found---should become less important as training progresses.
Hence, for each of our six models detailed in \cref{sec:methodology:models}, we consider the embeddings obtained at intervals of 1000 updates: i.e., we compute output decoder embeddings after $1000, 2000, \dots ,1000N$ updates. 
We can then measure whether scalar indicators defined in \cref{sec:methodology:metrics} differ across updates when using forced inference or beam-search.

In other words, we define series of paired scalar observations for each model, depending on which decomposition ($\mathrm{Dcp} \in \{\mathrm{Dcp}_{\mathrm{tok}}, \mathrm{Dcp}_{\mathrm{sl}}\}$), term (viz., $\mathbf{z} \in \{ \mathbf{i}, \mathbf{s}, \mathbf{t}, \mathbf{f}, \mathbf{c}\}$ for $\mathrm{Dcp}_{\mathrm{sl}}$ or $\mathbf{z} \in \{ \mathbf{c}, \mathbf{s}, \mathbf{t}\}$ for $\mathrm{Dcp}_{\mathrm{tok}}$) and indicator used ($f \in \{\mathrm{nr}, \cos\}$).
We pair, checkpoint by checkpoint, the average of the scalar indicator $f$ across our held-out test set when using either beam-search or forced inference, before computing correlation measures.

Remarkably, we find both Spearman's $\rho$ and Pearson's $r$ to be very highly correlated ($\rho > 0.986$ and $r > 0.901$) in all cases that we test.\footnote{
    Only 7 setups yield Pearson correlation coefficient below 0.99, all but one involving the $\mathbf{i}$ term of the $\mathrm{Dcp}_\mathrm{sl}$ decomposition: both cosine ($r > 0.944$) and norm-ratio  ($r > 0.977$) for the Indo-European-to-English model; the norm-ratio of the multilingual-to-English model ($r > 0.954$); and the cosine for the three Russian-to-English models (with $r > 0.901$, $r > 0.974$ and $r > 0.979$); the lowest of these Russian models also yield $r > 0.968$ for the $\mathbf{t}$ term  in $\mathrm{Dcp}_\mathrm{tok}$.
}
This extreme correlation indicates that, across training, embeddings derived through beam-search and embeddings derived through forced inference always exhibit the same geometric structures:
For instance, if for a given checkpoint, decomposition and term, we observe a low average cosine average between the said terms and the full embeddings as obtained through beam-search, then we are almost certain to obtain a similarly low cosine with forced inference as well.
In other words, corpus-level scalar indicators derived from linear decompositions do not appear to be sensitive to which decoding algorithm is used to compute embeddings.

\paragraph{Is geometry indicative of model performance at the corpus level?}

We have just established that linear decompositions appear stable across different means of decoding. 
This is broadly compatible with two interpretations:
either linear decompositions only capture idiosyncrasies of Transformer geometries; or there are other factors that could influence our scalar indicators.
One likely candidate would be model performances: 
We expect the embeddings of a highly performing model to differ significantly from that of a randomly initialized one or an under-trained one. By extension, differences in quality, as measured through automated metrics, should entail differences in geometry and in scalar indicators derived thereof.

For simplicity, let $\bar{f}(M)$ denote the average of applying function $f$ across our held-out dataset $\mathcal{D}$ using model $M$, i.e. 
$$ \bar{f}(M) = \frac{1}{|\mathcal{D}|}\sum_{x \in \mathcal{D}} f(M(x)) $$ 
To assess whether differences in geometry and quality are commensurate, we: 
\begin{itemize}
    \item[i)] sample pairs of models $M_{i}, M_{i+1}$
    \item[ii)] compute differences in scalar indicators $\bar{f}_\mathbf{z}(M_{i}) - \bar{f}_\mathbf{z}(M_{i+1})$, for $f_\mathbf{z}\in\{\mathrm{nr},\cos\}$ from \cref{eq:nr,eq:cos}; 
    \item[iii)] compute $\bar{f}_s(M_{i}) - \bar{f}_s(M_{i+1})$ for some scoring function $f_s$ such as BLEU;
    \item[iv)] compute the absolute value of Spearman correlation between these two series $\left| \rho \left( {S_{f_\mathbf{z}},S_{f_s}}^{} \right) \right|$
    \footnote{
    This is similar to performing a representational similarity analysis \citep{10.3389/neuro.06.004.2008} with the exception that we are looking at signed differences and computing the magnitude of the (anti-) correlations.
    Our aim is to capture whether scalar indicators and scoring functions are consistent with one another rather than determine what the optimal geometry is.
    As such, the directionality of a given effect is irrelevant (i.e., we do not care whether the cosine for a specific term has to be low or high for the model to perform well). 
}
\end{itemize}

\begin{figure}
    \centering
    \subfloat[$\mathrm{Dcp}_\mathrm{sl}$, $\cos$ and BLEU]{
    \begin{tikzpicture}
    \tikzstyle{every node}=[font=\tiny]
    \begin{axis}[
        ybar, ymin=0, ymax=100,
        symbolic x coords={I,S,T,F,C},
        bar width=.1cm,
        xtick=data,
        xticklabels={$\mathbf{i}$,$\mathbf{s}$,$\mathbf{t}$,$\mathbf{f}$,$\mathbf{c}$},
        enlarge x limits=0.125,
        enlarge y limits={upper=0.6},
        legend style={at={(0.5,1.05)},
            anchor=south,legend columns=-1,
            /tikz/every even column/.append style={column sep=0.5cm}
        },
        nodes near coords always on top/.style={
            scatter/position=absolute,
            positive value/.style={
                at={(axis cs:\pgfkeysvalueof{/data point/x},\pgfkeysvalueof{/data point/y})},
            },
            negative value/.style={
                at={(axis cs:\pgfkeysvalueof{/data point/x},0)},
            },
            every node near coord/.append style={
                check values/.code={%
                    \begingroup
                    \pgfkeys{/pgf/fpu}%
                    \pgfmathparse{\pgfplotspointmeta<0}%
                    \global\let\result=\pgfmathresult
                    \endgroup
                    %
                    %
                    \pgfmathfloatcreate{1}{1.0}{0}%
                    \let\ONE=\pgfmathresult
                    \ifx\result\ONE
                        \pgfkeysalso{/pgfplots/negative value}%
                    \else
                        \pgfkeysalso{/pgfplots/positive value}%
                    \fi
                },
                check values,
                anchor=west,
                rotate=90,
                font=\tiny,
                /pgf/number format/fixed,
                /pgf/number format/zerofill,
                /pgf/number format/precision=1,
                xshift=-0.5ex,
                color=black,
            },
        },
        nodes near coords={
            \pgfmathprintnumber[fixed zerofill,precision=1]{\pgfplotspointmeta}
        },
        nodes near coords always on top,
        height=3.5cm,
        width=1.1\columnwidth,
        xtick align=inside,
        label style={font=\tiny},
        cycle list/Blues-6,
        every axis plot/.append style={
            fill,
        },
    ]

 \addplot+ [draw=black] plot[error bars/.cd, y dir=both, y explicit] coordinates {
(I, 48.4771631971632)
(S, 71.4528416928417)
(T, 34.830231030231026)
(F, 46.02205758205758)
(C, 24.71114615114615)
};

 \addplot+ [draw=black] plot[error bars/.cd, y explicit] coordinates {
(I, 33.62628146628146)
(S, 68.31236499236498)
(T, 36.72183636183636)
(F, 56.2639015039015)
(C, 9.46199482199482)
};

 \addplot+ [draw=black] plot[error bars/.cd, y explicit] coordinates {
(I, 16.75552951552951)
(S, 72.11676803676804)
(T, 20.4187696987697)
(F, 55.02706830706832)
(C, 41.304460344460345)
};

 \addplot+ [draw=black] plot[error bars/.cd, y explicit] coordinates {
(I, 9.94182154182154)
(S, 50.93639609639611)
(T, 7.095589695589689)
(F, 53.28746664746665)
(C, 15.75144411144411)
};

 \addplot+ [draw=black] plot[error bars/.cd, y explicit] coordinates {
(I, 26.04411876411875)
(S, 64.1963003963004)
(T, 0.67180219180219)
(F, 71.97665985665986)
(C, 18.66331878331878)
};

 \addplot+ [draw=black] plot[error bars/.cd, y explicit] coordinates {
(I, 10.83388107388107)
(S, 68.72725868725868)
(T, 26.099274659274663)
(F, 68.12686460686461)
(C, 42.50383718383718)
};

\legend{\tt s0,\tt s1,\tt s2,\tt sla,\tt ine,\tt mul}
    \end{axis}
\end{tikzpicture}
    }
    
    \subfloat[$\mathrm{Dcp}_\mathrm{sl}$, $\mathrm{nr}$ and BLEU]{
    \begin{tikzpicture}
    \tikzstyle{every node}=[font=\tiny]
    \begin{axis}[
        ybar, ymin=0, ymax=100,
        symbolic x coords={I,S,T,F,C},
        bar width=.1cm,
        xtick=data,
        xticklabels={$\mathbf{i}$,$\mathbf{s}$,$\mathbf{t}$,$\mathbf{f}$,$\mathbf{c}$},
        enlarge x limits=0.125,
        enlarge y limits={upper=0.6},
        legend style={at={(0.5,1.05)},
            anchor=south,legend columns=-1,
            /tikz/every even column/.append style={column sep=0.5cm}
        },
        nodes near coords always on top/.style={
            scatter/position=absolute,
            positive value/.style={
                at={(axis cs:\pgfkeysvalueof{/data point/x},\pgfkeysvalueof{/data point/y})},
            },
            negative value/.style={
                at={(axis cs:\pgfkeysvalueof{/data point/x},0)},
            },
            every node near coord/.append style={
                check values/.code={%
                    \begingroup
                    \pgfkeys{/pgf/fpu}%
                    \pgfmathparse{\pgfplotspointmeta<0}%
                    \global\let\result=\pgfmathresult
                    \endgroup
                    %
                    %
                    \pgfmathfloatcreate{1}{1.0}{0}%
                    \let\ONE=\pgfmathresult
                    \ifx\result\ONE
                        \pgfkeysalso{/pgfplots/negative value}%
                    \else
                        \pgfkeysalso{/pgfplots/positive value}%
                    \fi
                },
                check values,
                anchor=west,
                rotate=90,
                font=\tiny,
                /pgf/number format/fixed,
                /pgf/number format/zerofill,
                /pgf/number format/precision=1,
                xshift=-0.5ex,
                color=black,
            },
        },
        nodes near coords={
            \pgfmathprintnumber[fixed zerofill,precision=1]{\pgfplotspointmeta}
        },
        nodes near coords always on top,
        height=3.5cm,
        width=1.1\columnwidth,
        xtick align=inside,
        label style={font=\tiny},
        cycle list/Blues-6,
        every axis plot/.append style={
            fill,
        },
    ]

 \addplot+ [draw=black] plot[error bars/.cd, y dir=both, y explicit] coordinates {
(I, 59.49625881625881)
(S, 57.02635274635275)
(T, 74.34404562404562)
(F, 57.61461925461926)
(C, 50.396645396645404)
};

 \addplot+ [draw=black] plot[error bars/.cd, y explicit] coordinates {
(I, 68.15610491610494)
(S, 54.46448158448158)
(T, 80.13221529221528)
(F, 31.93634473634473)
(C, 59.06509634509634)
};

 \addplot+ [draw=black] plot[error bars/.cd, y explicit] coordinates {
(I, 67.48569544569546)
(S, 28.47331491331492)
(T, 75.4184432984433)
(F, 56.63033471033471)
(C, 58.40106752106753)
};

 \addplot+ [draw=black] plot[error bars/.cd, y explicit] coordinates {
(I, 55.932851652851646)
(S, 28.904737784737776)
(T, 65.36133476133476)
(F, 1.8869573669573598)
(C, 65.05478317478317)
};

 \addplot+ [draw=black] plot[error bars/.cd, y explicit] coordinates {
(I, 49.21092913092913)
(S, 70.04781188781187)
(T, 75.3250712050712)
(F, 4.05544989544989)
(C, 64.92337608337608)
};

 \addplot+ [draw=black] plot[error bars/.cd, y explicit] coordinates {
(I, 25.66406866406867)
(S, 58.108406428406425)
(T, 45.9262094062094)
(F, 37.86342234342235)
(C, 72.81978585978585)
};

    \end{axis}
\end{tikzpicture}
    }
    
    \subfloat[$\mathrm{Dcp}_\mathrm{tok}$, $\cos$ and BLEU]{
        \resizebox{0.525\linewidth}{!}{
            \begin{tikzpicture}
    \tikzstyle{every node}=[font=\tiny]
    \begin{axis}[
        ybar, ymin=0, ymax=100,
        symbolic x coords={S,T,C},
        bar width=.1cm,
        xtick=data,
        xticklabels={$\mathbf{s}$,$\mathbf{t}$,$\mathbf{c}$},
        enlarge x limits=0.25,
        enlarge y limits={upper=0.6},
        legend style={at={(0.5,1.05)},
            anchor=south,legend columns=-1,
            /tikz/every even column/.append style={column sep=0.5cm}
        },
        nodes near coords always on top/.style={
            scatter/position=absolute,
            positive value/.style={
                at={(axis cs:\pgfkeysvalueof{/data point/x},\pgfkeysvalueof{/data point/y})},
            },
            negative value/.style={
                at={(axis cs:\pgfkeysvalueof{/data point/x},0)},
            },
            every node near coord/.append style={
                check values/.code={%
                    \begingroup
                    \pgfkeys{/pgf/fpu}%
                    \pgfmathparse{\pgfplotspointmeta<0}%
                    \global\let\result=\pgfmathresult
                    \endgroup
                    %
                    %
                    \pgfmathfloatcreate{1}{1.0}{0}%
                    \let\ONE=\pgfmathresult
                    \ifx\result\ONE
                        \pgfkeysalso{/pgfplots/negative value}%
                    \else
                        \pgfkeysalso{/pgfplots/positive value}%
                    \fi
                },
                check values,
                anchor=west,
                rotate=90,
                font=\tiny,
                /pgf/number format/fixed,
                /pgf/number format/zerofill,
                /pgf/number format/precision=1,
                xshift=-0.5ex,
                color=black,
            },
        },
        nodes near coords={
            \pgfmathprintnumber[fixed zerofill,precision=1]{\pgfplotspointmeta}
        },
        nodes near coords always on top,
        height=3.5cm,
        width=0.65\columnwidth,
        xtick align=inside,
        label style={font=\tiny},
        cycle list/Blues-6,
        every axis plot/.append style={
            fill,
        },
    ]

 \addplot+ [draw=black] plot[error bars/.cd, y dir=both, y explicit] coordinates {
(S, 26.875376875376876)
(T, 2.98998526998527)
(C, 38.725525645525636)
};

 \addplot+ [draw=black] plot[error bars/.cd, y explicit] coordinates {
(S, 31.717905997905998)
(T, 16.90351942351942)
(C, 6.574738534738531)
};

 \addplot+ [draw=black] plot[error bars/.cd, y explicit] coordinates {
(S, 33.01747165747166)
(T, 36.35682755682755)
(C, 23.48870888870889)
};

 \addplot+ [draw=black] plot[error bars/.cd, y explicit] coordinates {
(S, 27.30877254877255)
(T, 25.92595308595308)
(C, 48.83671067671066)
};

 \addplot+ [draw=black] plot[error bars/.cd, y explicit] coordinates {
(S, 8.97026397026397)
(T, 48.56213588213587)
(C, 2.19764427764427)
};

 \addplot+ [draw=black] plot[error bars/.cd, y explicit] coordinates {
(S, 23.92688056688056)
(T, 49.936740976740964)
(C, 22.545539985539982)
};

    \end{axis}
\end{tikzpicture}
        }
    }\hspace{-0.5cm}
    \subfloat[$\mathrm{Dcp}_\mathrm{tok}$, $\mathrm{nr}$ and BLEU]{
        \resizebox{0.475\linewidth}{!}{
            \begin{tikzpicture}
    \tikzstyle{every node}=[font=\tiny]
    \begin{axis}[
        ybar, ymin=0, ymax=100,
        symbolic x coords={S,T,C},
        bar width=.1cm,
        xtick=data,
        yticklabels={},
        xticklabels={$\mathbf{s}$,$\mathbf{t}$,$\mathbf{c}$},
        enlarge x limits=0.25,
        enlarge y limits={upper=0.6},
        legend style={at={(0.5,1.05)},
            anchor=south,legend columns=-1,
            /tikz/every even column/.append style={column sep=0.5cm}
        },
        nodes near coords always on top/.style={
            scatter/position=absolute,
            positive value/.style={
                at={(axis cs:\pgfkeysvalueof{/data point/x},\pgfkeysvalueof{/data point/y})},
            },
            negative value/.style={
                at={(axis cs:\pgfkeysvalueof{/data point/x},0)},
            },
            every node near coord/.append style={
                check values/.code={%
                    \begingroup
                    \pgfkeys{/pgf/fpu}%
                    \pgfmathparse{\pgfplotspointmeta<0}%
                    \global\let\result=\pgfmathresult
                    \endgroup
                    %
                    %
                    \pgfmathfloatcreate{1}{1.0}{0}%
                    \let\ONE=\pgfmathresult
                    \ifx\result\ONE
                        \pgfkeysalso{/pgfplots/negative value}%
                    \else
                        \pgfkeysalso{/pgfplots/positive value}%
                    \fi
                },
                check values,
                anchor=west,
                rotate=90,
                font=\tiny,
                /pgf/number format/fixed,
                /pgf/number format/zerofill,
                /pgf/number format/precision=1,
                xshift=-0.5ex,
                color=black,
            },
        },
        nodes near coords={
            \pgfmathprintnumber[fixed zerofill,precision=1]{\pgfplotspointmeta}
        },
        nodes near coords always on top,
        height=3.5cm,
        width=0.65\columnwidth,
        xtick align=inside,
        label style={font=\tiny},
        cycle list/Blues-6,
        every axis plot/.append style={
            fill,
        },
    ]

 \addplot+ [draw=black] plot[error bars/.cd, y dir=both, y explicit] coordinates {
(S, 29.67052419052419)
(T, 63.64711876711877)
(C, 63.405433605433615)
};

 \addplot+ [draw=black] plot[error bars/.cd, y explicit] coordinates {
(S, 15.68259308259308)
(T, 46.43078579078579)
(C, 61.973173013173025)
};

 \addplot+ [draw=black] plot[error bars/.cd, y explicit] coordinates {
(S, 43.55078231078231)
(T, 53.86297462297461)
(C, 56.00224052224052)
};

 \addplot+ [draw=black] plot[error bars/.cd, y explicit] coordinates {
(S, 24.889854049854048)
(T, 26.68581196581196)
(C, 31.25323757323757)
};

 \addplot+ [draw=black] plot[error bars/.cd, y explicit] coordinates {
(S, 66.6040134040134)
(T, 68.55457083457084)
(C, 61.90135402135402)
};

 \addplot+ [draw=black] plot[error bars/.cd, y explicit] coordinates {
(S, 26.91541647541647)
(T, 35.622391422391416)
(C, 43.78227442227443)
};

    \end{axis}
\end{tikzpicture}
        }
    }
    \caption{Corpus-level correlation magnitudes (Spearman's $|\rho|$, in \%) between scalar indicators ($\cos$, $\mathrm{nr}$) and BLEU. Remark the high variability across models, hinting at a lack of systematicity.}
    \label{fig:corpus-perf}
\end{figure}

We experiment with BLEU, COMET and chrF++ \citep{papineni-etal-2002-bleu,rei-etal-2020-comet,popovic-2017-chrf} as scoring functions. 
Corresponding results for BLEU are presented in \cref{fig:corpus-perf}.
We defer results with COMET and chrF++ to  \cref{adx:sup-res:perf-corpus}, \cref{fig:corpus-perf:extra:comet,fig:corpus-perf:extra:chrf++}, as they are in line with BLEU.
The notations \texttt{s0}, \texttt{s1} and \texttt{s2} refer to our three different runs for Russian-to-English; \texttt{sla}, \texttt{ine} and \texttt{mul} correspond to the Slavic-to-English, Indo-European-to-English and multilingual-to-English model.

There are several trends that we can observe.
First, correlation magnitudes tend to be high: 
This indicates that, on the whole, scalar indicators derived from linear decompositions tend to reflect model quality well (as captured by automatic metrics such as BLEU).
Second, and perhaps most importantly, we remark that results across the three seeds for Russian-to-English can display a high degree of variation, both in $\mathrm{Dcp}_\mathrm{tok}$ and $\mathrm{Dcp}_\mathrm{sl}$---for instance, in $\mathrm{Dcp}_\mathrm{sl}$, correlations between changes in cosines and changes in BLEU range from $|\rho|=9.5$ (in \texttt{s1}) to $|\rho|=41.3$ (\texttt{s2}) for the $\mathbf{c}$ term.
Third and last, the two decomposition approaches $\mathrm{Dcp}_\mathrm{tok}$ and $\mathrm{Dcp}_\mathrm{sl}$ suggest different interpretations as to how the decoders behave. 
For instance, compare target-side input tokens ($\mathbf{t}$ in  $\mathrm{Dcp}_\mathrm{tok}$) and target-side self-attention sub-layer outputs ($\mathbf{t}$ in  $\mathrm{Dcp}_\mathrm{sl}$):
While both terms aim to explain how the target-side input relates to the output embedding, the correlations we derive from the scalar indicators differ between decompositions. 
We find a surprisingly small correlation magnitude between $\cos$ and BLEU for the $\mathbf{t}$ term under $\mathrm{Dcp}_\mathrm{sl}$ in the \texttt{ine} model whereas the \texttt{s0} model presents the second highest magnitude---but turning to the same measurements for the $\mathbf{t}$ term under $\mathrm{Dcp}_\mathrm{tok}$, we find the exact opposite situation, with \texttt{s0} being noteworthily lower than all other models, and \texttt{ine} being the second highest.

In sum, while embedding geometry seems to be shaped in part by how effective a model is---as attested by the often high correlation scores we can observe---the relation between the two is neither straightforward nor systematic across models.

\paragraph{Is geometry indicative of model performance at the sentence level?}

We have thus far focused on corpus-level measurements. 
To test whether embedding geometry can provide useful explanations for specific inputs, it is important that we verify whether our observations also hold at the sentence level.

To broach this question, we consider the following methodology:
We first select a subset of $k=3000$ sentences; then for each sentence in said subset, we randomly select two checkpoints per seed.
We then compute the correlation magnitude between the signed differences in COMET scores and the signed differences in scalar indicators.
\footnote{
    We only focus on COMET as it has been suggested to be more appropriate for sentence-level quality estimation.
    We also conduct supplementary experiments in \cref{adx:sup-res:perf-sentence} with a slight modification of this methodology.
}

\begin{figure}
    \centering
    \subfloat[$\mathrm{Dcp}_\mathrm{sl}$, $\cos$ and COMET]{
    \begin{tikzpicture}
    \tikzstyle{every node}=[font=\tiny]
    \begin{axis}[
        ybar, ymin=0, ymax=100,
        symbolic x coords={I,S,T,F,C},
        bar width=.1cm,
        xtick=data,
        xticklabels={$\mathbf{i}$,$\mathbf{s}$,$\mathbf{t}$,$\mathbf{f}$,$\mathbf{c}$},
        enlarge x limits=0.125,
        enlarge y limits={upper=0.6},
        legend style={at={(0.5,1.05)},
            anchor=south,legend columns=-1,
            /tikz/every even column/.append style={column sep=0.5cm}
        },
        nodes near coords always on top/.style={
            scatter/position=absolute,
            positive value/.style={
                at={(axis cs:\pgfkeysvalueof{/data point/x},\pgfkeysvalueof{/data point/y})},
            },
            negative value/.style={
                at={(axis cs:\pgfkeysvalueof{/data point/x},0)},
            },
            every node near coord/.append style={
                check values/.code={%
                    \begingroup
                    \pgfkeys{/pgf/fpu}%
                    \pgfmathparse{\pgfplotspointmeta<0}%
                    \global\let\result=\pgfmathresult
                    \endgroup
                    %
                    %
                    \pgfmathfloatcreate{1}{1.0}{0}%
                    \let\ONE=\pgfmathresult
                    \ifx\result\ONE
                        \pgfkeysalso{/pgfplots/negative value}%
                    \else
                        \pgfkeysalso{/pgfplots/positive value}%
                    \fi
                },
                check values,
                anchor=west,
                rotate=90,
                font=\tiny,
                /pgf/number format/fixed,
                /pgf/number format/zerofill,
                /pgf/number format/precision=1,
                xshift=-0.5ex,
                color=black,
            },
        },
        nodes near coords={
            \pgfmathprintnumber[fixed zerofill,precision=1]{\pgfplotspointmeta}
        },
        nodes near coords always on top,
        height=3.5cm,
        width=1.1\columnwidth,
        xtick align=inside,
        label style={font=\tiny},
        cycle list/Purples-6,
        every axis plot/.append style={
            fill,
        },
    ]

 \addplot+ [draw=black] plot[error bars/.cd, y dir=both, y explicit] coordinates {
	(I, 18.134207516785796 ) 
	(S, 17.735942603718804 ) 
	(T, 7.4944852695733335 ) 
	(F, 10.85097670180251 ) 
	(C, 6.354912727772676 ) 
};

 \addplot+ [draw=black] plot[error bars/.cd, y explicit] coordinates {
	(I, 9.123944879482137 ) 
	(S, 16.29229278607195 ) 
	(T, 5.968017813082031 ) 
	(F, 15.622769927413078 ) 
	(C, 3.510037796341531 ) 

};

 \addplot+ [draw=black] plot[error bars/.cd, y explicit] coordinates {
	(I, 12.653411529982685 ) 
	(S, 13.883105572968093 ) 
	(T, 3.510672089743503 ) 
	(F, 10.026469609682474 ) 
	(C, 4.838911108948696 ) 
};

 \addplot+ [draw=black] plot[error bars/.cd, y explicit] coordinates {
	(I, 8.230834148801632 ) 
	(S, 19.92518253139335 ) 
	(T, 5.1870332925235125 ) 
	(F, 15.294349158359866 ) 
	(C, 2.174159344746176 ) 

};

 \addplot+ [draw=black] plot[error bars/.cd, y explicit] coordinates {
	(I, 7.924625169284301 ) 
	(S, 26.019891085370734 ) 
	(T, 3.1646661337580033 ) 
	(F, 24.613296033392576 ) 
	(C, 4.449815724274047 ) 

};

 \addplot+ [draw=black] plot[error bars/.cd, y explicit] coordinates {
	(I, 10.503700804856708 ) 
	(S, 24.97176121001507 ) 
	(T, 5.572465350117434 ) 
	(F, 26.66836327319549 ) 
	(C, 5.899119952206979 ) 

};

\legend{\tt s0,\tt s1,\tt s2,\tt sla,\tt ine,\tt mul}
    \end{axis}
\end{tikzpicture}
    }
    
    \subfloat[$\mathrm{Dcp}_\mathrm{sl}$, $\mathrm{nr}$ and COMET]{
    \begin{tikzpicture}
    \tikzstyle{every node}=[font=\tiny]
    \begin{axis}[
        ybar, ymin=0, ymax=100,
        symbolic x coords={I,S,T,F,C},
        bar width=.1cm,
        xtick=data,
        xticklabels={$\mathbf{i}$,$\mathbf{s}$,$\mathbf{t}$,$\mathbf{f}$,$\mathbf{c}$},
        enlarge x limits=0.125,
        enlarge y limits={upper=0.6},
        legend style={at={(0.5,1.05)},
            anchor=south,legend columns=-1,
            /tikz/every even column/.append style={column sep=0.5cm}
        },
        nodes near coords always on top/.style={
            scatter/position=absolute,
            positive value/.style={
                at={(axis cs:\pgfkeysvalueof{/data point/x},\pgfkeysvalueof{/data point/y})},
            },
            negative value/.style={
                at={(axis cs:\pgfkeysvalueof{/data point/x},0)},
            },
            every node near coord/.append style={
                check values/.code={%
                    \begingroup
                    \pgfkeys{/pgf/fpu}%
                    \pgfmathparse{\pgfplotspointmeta<0}%
                    \global\let\result=\pgfmathresult
                    \endgroup
                    %
                    %
                    \pgfmathfloatcreate{1}{1.0}{0}%
                    \let\ONE=\pgfmathresult
                    \ifx\result\ONE
                        \pgfkeysalso{/pgfplots/negative value}%
                    \else
                        \pgfkeysalso{/pgfplots/positive value}%
                    \fi
                },
                check values,
                anchor=west,
                rotate=90,
                font=\tiny,
                /pgf/number format/fixed,
                /pgf/number format/zerofill,
                /pgf/number format/precision=1,
                xshift=-0.5ex,
                color=black,
            },
        },
        nodes near coords={
            \pgfmathprintnumber[fixed zerofill,precision=1]{\pgfplotspointmeta}
        },
        nodes near coords always on top,
        height=3.5cm,
        width=1.1\columnwidth,
        xtick align=inside,
        label style={font=\tiny},
        cycle list/Purples-6,
        every axis plot/.append style={
            fill,
        },
    ]

 \addplot+ [draw=black] plot[error bars/.cd, y dir=both, y explicit] coordinates {
	(I, 8.671943011911905 ) 
	(S, 15.60873099277218 ) 
	(T, 18.531590674568722 ) 
	(F, 15.738643767214198 ) 
	(C, 15.48349562086129 ) 

};

 \addplot+ [draw=black] plot[error bars/.cd, y explicit] coordinates {
	(I, 14.004577573110161 ) 
	(S, 15.696243308380994 ) 
	(T, 19.694927615380077 ) 
	(F, 8.989359356824867 ) 
	(C, 16.74606249097037 ) 

};

 \addplot+ [draw=black] plot[error bars/.cd, y explicit] coordinates {
	(I, 6.986469275764749 ) 
	(S, 12.311156075466398 ) 
	(T, 13.826477706228491 ) 
	(F, 8.423529008489524 ) 
	(C, 13.81721291477665 ) 

};

 \addplot+ [draw=black] plot[error bars/.cd, y explicit] coordinates {
	(I, 11.692297246003 ) 
	(S, 15.607407348189154 ) 
	(T, 18.64767769386792 ) 
	(F, 6.874516901892683 ) 
	(C, 21.51563674977432 ) 

};

 \addplot+ [draw=black] plot[error bars/.cd, y explicit] coordinates {
	(I, 10.292767499882133 ) 
	(S, 24.497330374997492 ) 
	(T, 25.91110324093729 ) 
	(F, 4.977838967393022 ) 
	(C, 26.337506688189237 ) 

};

 \addplot+ [draw=black] plot[error bars/.cd, y explicit] coordinates {
	(I, 5.265900360112027 ) 
	(S, 17.606218425639845 ) 
	(T, 27.431725387224077 ) 
	(F, 8.47906454057821 ) 
	(C, 29.887025554675734 ) 

};

    \end{axis}
\end{tikzpicture}
    }
    
    \subfloat[$\mathrm{Dcp}_\mathrm{tok}$, $\cos$ and COMET]{
        \resizebox{0.525\linewidth}{!}{
            \begin{tikzpicture}
    \tikzstyle{every node}=[font=\tiny]
    \begin{axis}[
        ybar, ymin=0, ymax=100,
        symbolic x coords={S,T,C},
        bar width=.1cm,
        xtick=data,
        xticklabels={$\mathbf{s}$,$\mathbf{t}$,$\mathbf{c}$},
        enlarge x limits=0.25,
        enlarge y limits={upper=0.6},
        legend style={at={(0.5,1.05)},
            anchor=south,legend columns=-1,
            /tikz/every even column/.append style={column sep=0.5cm}
        },
        nodes near coords always on top/.style={
            scatter/position=absolute,
            positive value/.style={
                at={(axis cs:\pgfkeysvalueof{/data point/x},\pgfkeysvalueof{/data point/y})},
            },
            negative value/.style={
                at={(axis cs:\pgfkeysvalueof{/data point/x},0)},
            },
            every node near coord/.append style={
                check values/.code={%
                    \begingroup
                    \pgfkeys{/pgf/fpu}%
                    \pgfmathparse{\pgfplotspointmeta<0}%
                    \global\let\result=\pgfmathresult
                    \endgroup
                    %
                    %
                    \pgfmathfloatcreate{1}{1.0}{0}%
                    \let\ONE=\pgfmathresult
                    \ifx\result\ONE
                        \pgfkeysalso{/pgfplots/negative value}%
                    \else
                        \pgfkeysalso{/pgfplots/positive value}%
                    \fi
                },
                check values,
                anchor=west,
                rotate=90,
                font=\tiny,
                /pgf/number format/fixed,
                /pgf/number format/zerofill,
                /pgf/number format/precision=1,
                xshift=-0.5ex,
                color=black,
            },
        },
        nodes near coords={
            \pgfmathprintnumber[fixed zerofill,precision=1]{\pgfplotspointmeta}
        },
        nodes near coords always on top,
        height=3.5cm,
        width=0.65\columnwidth,
        xtick align=inside,
        label style={font=\tiny},
        cycle list/Purples-6,
        every axis plot/.append style={
            fill,
        },
    ]

 \addplot+ [draw=black] plot[error bars/.cd, y dir=both, y explicit] coordinates {
	(S, 7.531107108349396 ) 
	(T, 8.051043497684848 ) 
	(C, 4.963457275030816 ) 

};

 \addplot+ [draw=black] plot[error bars/.cd, y explicit] coordinates {
	(S, 7.693835904912868 ) 
	(T, 2.401698260433449 ) 
	(C, 4.192268488189557 ) 

};

 \addplot+ [draw=black] plot[error bars/.cd, y explicit] coordinates {
	(S, 10.806960847060784 ) 
	(T, 0.49967582015144646 ) 
	(C, 6.251502441480534 ) 

};

 \addplot+ [draw=black] plot[error bars/.cd, y explicit] coordinates {
	(S, 13.733596772162244 ) 
	(T, 1.5977901802613381 ) 
	(C, 13.769710044985715 ) 
};

 \addplot+ [draw=black] plot[error bars/.cd, y explicit] coordinates {
	(S, 12.32654586962541 ) 
	(T, 10.560706328340725 ) 
	(C, 11.236813822635522 ) 

};

 \addplot+ [draw=black] plot[error bars/.cd, y explicit] coordinates {
	(S, 2.8399679407766834 ) 
	(T, 4.008583301150199 ) 
	(C, 17.70373364160866 ) 

};

    \end{axis}
\end{tikzpicture}
        }
    }\hspace{-0.5cm}
    \subfloat[$\mathrm{Dcp}_\mathrm{tok}$, $\mathrm{nr}$ and COMET]{
        \resizebox{0.475\linewidth}{!}{
            \begin{tikzpicture}
    \tikzstyle{every node}=[font=\tiny]
    \begin{axis}[
        ybar, ymin=0, ymax=100,
        symbolic x coords={S,T,C},
        bar width=.1cm,
        xtick=data,
        yticklabels={},
        xticklabels={$\mathbf{s}$,$\mathbf{t}$,$\mathbf{c}$},
        enlarge x limits=0.25,
        enlarge y limits={upper=0.6},
        legend style={at={(0.5,1.05)},
            anchor=south,legend columns=-1,
            /tikz/every even column/.append style={column sep=0.5cm}
        },
        nodes near coords always on top/.style={
            scatter/position=absolute,
            positive value/.style={
                at={(axis cs:\pgfkeysvalueof{/data point/x},\pgfkeysvalueof{/data point/y})},
            },
            negative value/.style={
                at={(axis cs:\pgfkeysvalueof{/data point/x},0)},
            },
            every node near coord/.append style={
                check values/.code={%
                    \begingroup
                    \pgfkeys{/pgf/fpu}%
                    \pgfmathparse{\pgfplotspointmeta<0}%
                    \global\let\result=\pgfmathresult
                    \endgroup
                    %
                    %
                    \pgfmathfloatcreate{1}{1.0}{0}%
                    \let\ONE=\pgfmathresult
                    \ifx\result\ONE
                        \pgfkeysalso{/pgfplots/negative value}%
                    \else
                        \pgfkeysalso{/pgfplots/positive value}%
                    \fi
                },
                check values,
                anchor=west,
                rotate=90,
                font=\tiny,
                /pgf/number format/fixed,
                /pgf/number format/zerofill,
                /pgf/number format/precision=1,
                xshift=-0.5ex,
                color=black,
            },
        },
        nodes near coords={
            \pgfmathprintnumber[fixed zerofill,precision=1]{\pgfplotspointmeta}
        },
        nodes near coords always on top,
        height=3.5cm,
        width=0.65\columnwidth,
        xtick align=inside,
        label style={font=\tiny},
        cycle list/Purples-6,
        every axis plot/.append style={
            fill,
        },
    ]

 \addplot+ [draw=black] plot[error bars/.cd, y dir=both, y explicit] coordinates {
	(S, 13.591027553469553 ) 
	(T, 15.718004179323625 ) 
	(C, 16.912807434943524 ) 

};

 \addplot+ [draw=black] plot[error bars/.cd, y explicit] coordinates {
	(S, 11.735008377591761 ) 
	(T, 14.792938963024294 ) 
	(C, 15.580412634820595 ) 

};

 \addplot+ [draw=black] plot[error bars/.cd, y explicit] coordinates {
	(S, 12.490700083103626 ) 
	(T, 15.523918526996935 ) 
	(C, 16.559839325807584 ) 

};

 \addplot+ [draw=black] plot[error bars/.cd, y explicit] coordinates {
	(S, 17.806338767663952 ) 
	(T, 22.589012724323716 ) 
	(C, 22.164203809403247 ) 

};

 \addplot+ [draw=black] plot[error bars/.cd, y explicit] coordinates {
	(S, 20.78487988435132 ) 
	(T, 25.831613718587022 ) 
	(C, 26.956239748054365 ) 

};

 \addplot+ [draw=black] plot[error bars/.cd, y explicit] coordinates {
	(S, 19.73861607719478 ) 
	(T, 22.477210229961045 ) 
	(C, 23.70324585117002 ) 

};

    \end{axis}
\end{tikzpicture}
        }
    }
    \caption{Sentence-level correlation magnitudes (Spearman's $|\rho|$, in \%) between scalar indicators ($\cos$, $\mathrm{nr}$) and COMET. Magnitudes are often much lower than their counterparts in \cref{fig:corpus-perf}, suggesting a poorer fit.}
    \label{fig:sentence-perf}
\end{figure}

Corresponding results are provided in \cref{fig:sentence-perf}.
We can make two important remarks.
First, we can see that correlation scores are often much lower than what we observed at the corpus level.\footnote{
    In fact the $p$-value provided by \texttt{scipy} for these correlation scores suggests that many of these correlations are spurious.
}
Nonetheless, some setups still perform reliably well---in particular, norm ratio is found to yield higher correlation magnitudes than cosine n $\mathrm{Dcp}_\mathrm{tok}$.
This would entail that model quality factors in the results we obtain at the sentence level---if to a lesser extent.
Second, we still observe important variation across all three seeds for Russian---often comparable to variation attested across training conditions.

This result suggests that geometry-based explanations are more in line with corpus-level statistics than with sentence-level observations.
This naturally questions their usefulness as far as model explainability is concerned, and echoes our previous findings about decoding algorithms:
We established that forced inference and beam search did not entail different geometries, we now observe that sentence-level quality is often less appropriate than corpus-level quality when attempting to account for the geometry a model settles on.

\paragraph{Is geometry indicative of training conditions?}
Throughout our previous experiments, we have seen that variation across our three Russian models was often comparable to variation across different training datasets. 
We now turn as to whether this fact can be established more firmly: 
Is there evidence that models that are trained in similar circumstances develop similar geometry?
One important aspect of this question consists in assessing the \emph{evolution across training}, rather than focusing on individual checkpoints as we have thus far.

Thus, we now consider the time-series described by our scalar indicators in \cref{eq:nr,eq:cos} for each term $\mathbf{z}$ of a given decomposition $\mathrm{Dcp}$, through the entire training. 
For each term and indicator, we compare the time series of all different models using the dynamic time warping algorithm \cite[DTW,][]{bellman59dtw,sakoe78dtw}. 
Our interest in doing this comparison resides in being able to understand how distant the time series of the different models are between them. 
The DTW algorithm is especially suitable to our use case, as it measures similarity in a manner that is invariant to shifts and length differences between two time series. 
In other words, it allows us to measure how similar the series 
$\mathrm{nr}_{M_1}(\mathbf{z},\mathbf{e})|_{1,...,N_1}$ and 
$\mathrm{nr}_{M_2}(\mathbf{z},\mathbf{e})|_{1,...,N_2}$ are, disregarding the different speed of convergence of both models $M_1$ and $M_2$ at training time.  

\begin{figure*}[ht]
    \centering
    \subfloat[\label{fig:dtw:tok} DTW distances for $\mathrm{Dcp}_\mathrm{tok}$]{
        \includegraphics[width=0.6\linewidth, trim={0 3.45cm 1cm 1.4cm},clip]{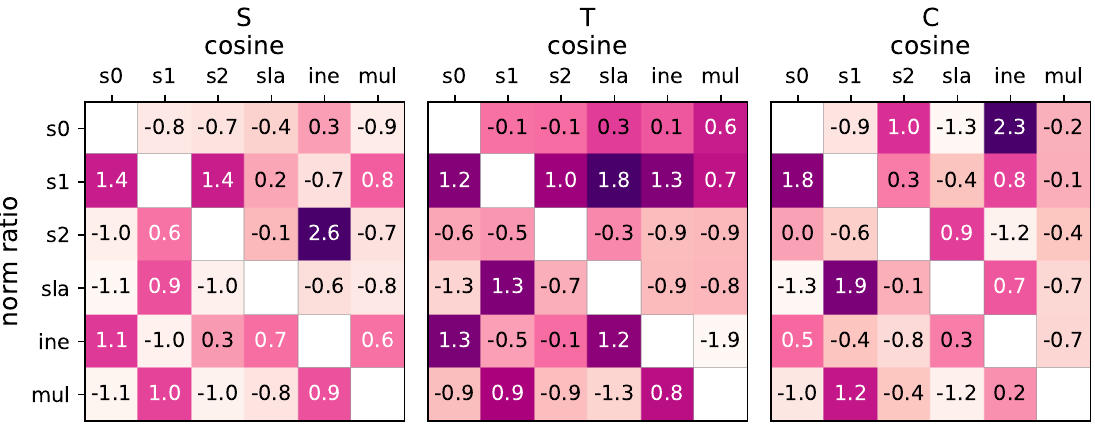}
    }
    
    \subfloat[\label{fig:dtw:sl} DTW distances for $\mathrm{Dcp}_\mathrm{sl}$]{
        \includegraphics[width=0.99\linewidth, trim={0 4.8cm 0.75cm 1.27cm},clip]{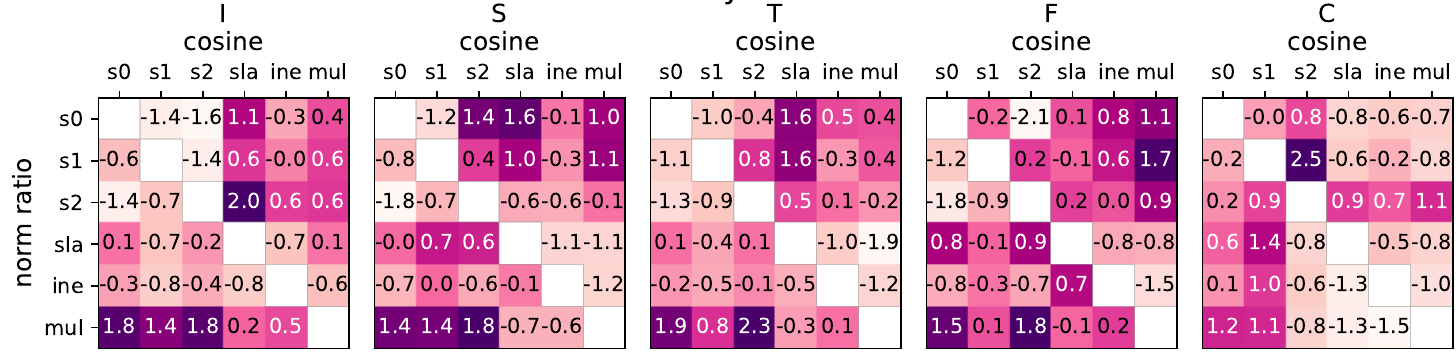}
    }
    \caption{Dynamic time warping distance measurements, $z$-normalized. Remark that distances between seed replications (\texttt{s0}, \texttt{s1}, \texttt{s0}) do not differ from distances between models with different inputs.}
    \label{fig:dtw}
\end{figure*}

Corresponding results are provided in \cref{fig:dtw}.
Each of the heatmap corresponds to the time-series relating to a given term. The upper triangle of each heatmap relates to cosine, and the lower triangle to norm-ratio time series.
Individual cells indicate the distance between the time series derived for the models listed in row and column.
For instance, the cell in row 2, column 4 of the third plot in \cref{fig:dtw:tok} corresponds to the distance  measured between cosine measurements of the $\mathbf{c}$ term under $\mathrm{Dcp}_\mathrm{tok}$ in the \texttt{s1} and \texttt{sla} models.
Results are $z$-normalized, as our interest lies in verifying whether Russian models are distinct from other models rather than establish the absolute difference.

The three Russian seeds correspond to the top three rows and columns in each heatmap.
A natural expectation would be that comparisons between Russian seeds should lead to more similar time series, and thus lower ($z$-normalized) distances.
Instead, what we observe is consistent with previous experiments:
Comparisons between two Russian seeds may or may not yield lower distances.
In particular, \texttt{s1} and \texttt{s2} often yield very distinct time-series, i.e., the models develop very different geometries despite their similar training conditions.

\begin{table}[ht]
    \centering
    \nprounddigits{3}
    \npdecimalsign{.}
    \subfloat[$\mathrm{Dcp}_\mathrm{sl}$]{
    \begin{tabular}{l n{1}{3} n{1}{3} n{1}{3} n{1}{3} n{1}{3}}
\toprule
& \multicolumn{5}{c}{\textbf{term}} \\
& {{$\mathbf{i}$}} &  {{$\mathbf{s}$}} &  {{$\mathbf{t}$}} &  {{$\mathbf{f}$}} &  {{$\mathbf{c}$}} \\
\midrule
$\cos$ & 0.002198 &  0.367033 & 0.351648  & 0.107692 & 0.021978 \\
$\mathrm{nr}$ & 0.019780 &  0.002198 & 0.002198  & 0.002198 & 0.296703 \\
\bottomrule
\end{tabular}
}

\subfloat[$\mathrm{Dcp}_\mathrm{tok}$]{
    \begin{tabular}{l n{1}{3} n{1}{3} n{1}{3}}
\toprule
& \multicolumn{3}{c}{\textbf{term}} \\
& {{$\mathbf{s}$}} &  {{$\mathbf{t}$}} &  {{$\mathbf{c}$}} \\
\midrule
$\cos$ & 0.503297 & 0.316484 & 0.380220 \\
$\mathrm{nr}$ & 0.221978 & 0.421978 & 0.230769 \\
\bottomrule
\end{tabular}
}
    \caption{$p$-values derived from Pitman permutation tests}
    \label{tab:pitman-dtw}
\end{table}

To provide a more thorough outlook on this question, we conduct Pitman permutation tests \citep{dror-etal-2018-hitchhikers} to establish whether comparisons between two Russian models are statistically lower than others.
Corresponding results are provided in \cref{tab:pitman-dtw}. 
As we can see, while select setups using $\mathrm{Dcp}_\mathrm{sl}$ yield $p$-values beyond the commonly used 0.05 threshold, only half of the setup we experiment with yield the expected result. 
In particular, all setups based on $\mathrm{Dcp}_\mathrm{tok}$ are insignificant.

We therefore conclude that different decomposition approaches lead to different interpretations of what Transformer geometry encodes.
Had we only focused on $\mathrm{Dcp}_\mathrm{tok}$, we would have been lead to a much firmer rejection of the notion that decompositions are stable across random initializations.
The inclusion of $\mathrm{Dcp}_\mathrm{sl}$ in our experiments forces us to adopt a more nuanced approach: viz., that the evidence in favor of geometry-based explainability approaches is thin; and that results derived from such approaches appear very brittle---the exact methodology used brings about variations in $p$-value of up to two orders of magnitude.

\section{Conclusions}

We have presented a series of statistical studies questioning the usefulness of linear decomposition approaches.
In particular, we have highlighted that straightforward vector space characteristics, such as angle and norm of the derived vector terms, imply the following three points: (i) decompositions are invariant to the decoding algorithm employed; (ii) they are more in line with corpus-level performance than sentence-level performance, and (iii) variance across random seeds for the same training conditions is often comparable to variance across models trained on different corpora.
Taken together, our experiments suggest that Transformer geometry is often highly model-specific. 
Observations about a specific model need not generalize.

As such, some of the assumptions underlying geometry-based explanations of Transformer behaviors are not borne out.
While it is true that the geometry of successful models differs from that of unsuccessful ones, our work puts forth evidence that this difference is mostly trivial---geometry being model-specific necessarily entails that any partition of models, be it based on performance or else, will naturally highlight differences.

While our focus has been limited to linear decompositions and straightforward vector characteristics, our experiments more broadly call into question the validity of many related approaches, which we hope to investigate in future work.
That straightforward vector characteristics do not yield a coherent picture \textsl{a minima} entails that linear decomposition approaches have to rely on non-straightforward, high-dimensional relationships. 
That similar training conditions cannot guarantee similar vector spaces naturally leads us to doubt the generalization power of methodologies that probe a handful of foundational models: 
If we are unable to ensure that our approaches would generalize to other similar models, can we truthfully say that the explanations we provide are indeed reasonable?

\section*{Acknowledgments}
We thank Hande Celikkanat, Denis Paperno and the three anonymous reviewers for their insightful comments, as well as Jörg Tiedemann for invaluable help with data selection. \\[-0.25em]

\noindent
{ 
\begin{minipage}{0.1\linewidth}
    \vspace{-10pt}
    \raisebox{-0.2\height}{\includegraphics[trim =32mm 55mm 30mm 5mm, clip, scale=0.18]{erc.ai}} \\[0.25cm]
    \raisebox{-0.25\height}{\includegraphics[trim =0mm 5mm 5mm 2mm,clip,scale=0.075]{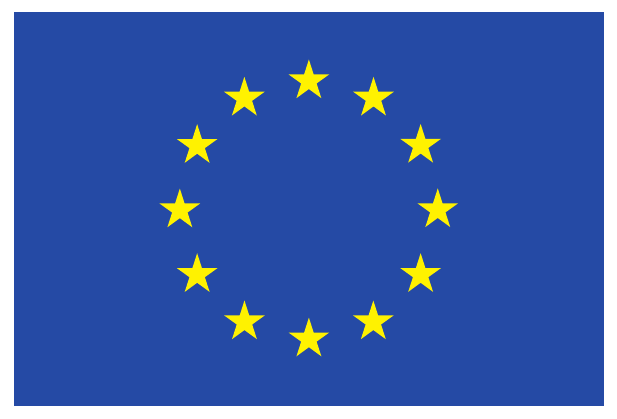}}
\end{minipage}
\hspace{0.01\linewidth}
\begin{minipage}{0.85\linewidth}
This work is part of the FoTran project, funded by the European Research Council (ERC) under the EU's Horizon 2020 research and innovation program (agreement \textnumero{}~771113). ~~We ~also ~thank ~the CSC-IT\vspace{0.3ex}
\end{minipage}
\begin{minipage}{\linewidth}
\noindent  Center for Science Ltd., for computational resources.
\end{minipage}%
}
\bibliography{anthology,custom}
\bibliographystyle{acl_natbib}

\appendix

\section{Model training details}
\label{adx:hparams}

As noted in the main text, we use the Tatoeba Challenge corpus \citep{tiedemann-2020-tatoeba} and the marian-MT library \citep{junczys-dowmunt-etal-2018-marian}.
Models were trained using four V100 nVidia GPUs.

Models use sources of different degrees of multilinguality: multilingual-to-English (\texttt{mul}); Indo-European-to-English (\texttt{ine}); Slavic-to-English (\texttt{sla}) and three different seeds for Russian-to-English (\texttt{s0}, \texttt{s1} and \texttt{s2}).
All languages included in a more specific model are also present in all more multilingual models.
For instance, there are datapoints in the \texttt{mul} model's training data for each of the Slavic languages used to train the \texttt{sla} model.
A complete list of the languages used for multilingual models in this study can be found in \cref{tab:langs}; all models also contain Russian.

\begin{table}[!t]
    \centering
    \begin{tabular}{>{\bf}m{0.025\linewidth} >{\footnotesize\ttfamily}m{0.85\linewidth}}
    \toprule
        \rotatebox{90}{Slavic} & bel bos bul ces cnr csb dsb hbs hrv hsb mkd pol slk slv srp szl ukr \\
    \midrule
        \rotatebox{90}{Indo-European}  & afr anp arg asm ast bar ben bis bre bzj cat ckb cor cos crs cym dan deu div djk dty ell fao fas fra frp fry fur gla gle glg glv guj hat hin hne hye hyw ind isl ita jak kas kea kmr kri kur lad lav lij lim lit lmo ltg ltz mai mar mfe min mol msa mwl nds nep nld nno nob oci ori oss pan pap pdt pes pis pob por prs pus rmn rmy roh rom ron san scn sco sin spa sqi srd srm srn swe tgk tpi urd vec wae wes wln yid zlm \\
    \midrule
        \rotatebox{90}{multilingual} & abk ace ach ada aka alt alz amh ami ara arq ary arz ava aze azz bak bas bbc bci bcl bem bhw bin bod brx bts btx bug bum cab cak ceb cha chk chr chv cjk cmn cnh cop crh  ctu dhv dik din dje dua dyu dzo efi epo est eus ewe fas fij fil fin fon ful fuv gil grn guc gug guw gym heb her hil hmn hne hun iba ibg ibo ido ilo ish iso ixl jav jbo jpn kab kac kal kam kan kat kau kaz kbp kek khm kik kin kmb kon koo kqn kss ksw kua kwn lam lao lfn lin loz lua lub lue lug lun luo lus lzh mah mal mam mau meh men mgr mhr mlg mlt mon mos mri mrj mxv mya nan naq nav nba nbl nch ncj ncx ndc nde ndo ngl ngu nia nij niu nso nya nyk nyn nyu nzi oke orm pag plt pon quc rar rnd run sag sat seh ses sid sme smo sna som sop sot ssw sun swa sxn syr tah tam tat tcf tdt tel tgl tha tir tiv tll tmh tog toh toi toj ton trs tsc tsn tso ttj tuk tum tur tvl twi tyv tzh tzo udm uig umb urh uzb ven vie vmw wal war wls wol wuu xho xmf yao yap yor yua yue zai zam zne zpa zul \\
    \bottomrule
    \end{tabular}
    \caption{List of language sources for multilingual models. More multilingual sources also contain languages from less multilingual models. All models also contain Russian.}
    \label{tab:langs}
\end{table}

Detailed hyperparameters is provided in \cref{tab:hparams}.
We refer the reader to \citet{junczys-dowmunt-etal-2018-marian} and the associated documentation\footnote{\url{https://marian-nmt.github.io/docs/}} for further explanations.
Models \texttt{s0}, \texttt{sla}, \texttt{ine} and \texttt{mul} used the first of the three listed seeds, whereas \texttt{s1} used the second and \texttt{s2} the third.
In practice, none of the six models fulfilled the early stopping criterion in the allocated runtime (72h).

\begin{table}[]
    \centering
    \begin{tabular}{>{\bf}m{0.4\columnwidth} >{\ttfamily}m{0.4\columnwidth}}
\toprule
\textbf{H-param.} & \textbf{\rmfamily Value} \\ \midrule
type & transformer \\
quiet-translation & true \\
max-length & 500 \\
mini-batch-fit & true \\
workspace & 24000 \\
maxi-batch & 500 \\
valid-mini-batch & 16 \\
valid-freq & 5000 \\
save-freq & 1000 \\
disp-freq & 5000 \\
valid-metrics & perplexity cross-entropy bleu chrf \\
beam-size & 12 \\
normalize & 1 \\
allow-unk & true \\
enc-depth & 6 \\
dec-depth & 6 \\
transformer-heads & 8 \\
transformer-postprocess-emb & d \\
transformer-postprocess & dan \\
transformer-ffn-activation & swish \\
transformer-dropout & 0.1 \\
label-smoothing & 0.1 \\
learn-rate & 0.0003 \\
lr-warmup & 16000 \\
lr-decay-inv-sqrt & 16000 \\
lr-report & true \\
optimizer-params & 0.9 0.98 1e-09 \\
clip-norm & 5 \\
fp16 & true \\
tied-embeddings-all & true \\
early-stopping & 150 \\
cost-type & ce-mean \\
exponential-smoothing & true \\
devices & 0 1 2 3  \\
sync-sgd & true \\
seed & 1111 1989 20232 \\
\bottomrule
    \end{tabular}
    \caption{\label{tab:hparams} Hyperparameters for models}
\end{table}

\section{Supplementary details on decompositions}
\label{adx:dcp-supinfo}

\paragraph{Notation details.}

\cref{tab:notations} lists the notations used throughout this work.
Remark that, aside from row-selection (marked $\left(\mathbf{Z}\right)_i$), symbol typesetting indicates the type of mathematical object denoted: i.e., $a_{\lambda htt'}$ is a scalar and not a tensor of rank 4.

\paragraph{Presentation of the Transformer decoder architecture.}
The remainder of this appendix consists in a general introduction to a Transformer decoder architecture.
We refer the reader to \citet{NIPS2017_3f5ee243} for a more thorough overview.

A Transformer decoder is a stack of $L$ layers, each containing 3 sub-layers.
Sub-layers are defined by means of specific sub-layer components: either \emph{multi-head attention mechanisms} ($\mathrm{ma}$) or \emph{feed-forwards} ($\mathrm{ff}$).

The latter are multi-layer perceptrons of the form:
\begin{equation*}
\resizebox{0.95\linewidth}{!}{$
    \mathbf{\dot{e}}_{\lambda, t} = \mathbf{W}^{(\mathrm{ff,out})}_\lambda \phi \left( \mathbf{W}^{(\mathrm{ff,in})}_\lambda \mathbf{x}_{\lambda,t} + \mathbf{b}^{(\mathrm{ff,in})}_\lambda \right) + \mathbf{b}^{(\mathrm{ff}, out)}_\lambda
$}
\end{equation*}
where $\phi$ is a non-linear activation function (e.g., $\mathrm{ReLU}$, $\mathrm{SiLU}$, $\mathrm{GELU}$...).

Multi-head attention mechanisms consist in attention-based weighted average computations:
\begin{equation*}
\resizebox{0.95\linewidth}{!}{$%
\begin{aligned}
    \mathbf{\dot{E}}_{\lambda} &= \mathbf{W}^{(\mathrm{ma,O})}_\lambda \bigoplus\limits_{h=1}^H \mathbf{A}_{\lambda,h} \left( \mathbf{W}^{(\mathrm{ma,V})}_{\lambda,h} \mathbf{X} + \mathbf{b}^{(\mathrm{ma,V})}_{\lambda,h} \right) \\ & \qquad + \mathbf{b}^{(\mathrm{ma,O})}_\lambda
\end{aligned}
$}%
\end{equation*}
where the input $\mathbf{X}$ is either the previous sub-layer output, up to token $t$ included (i.e., $\left[\begin{smallmatrix}
    \mathbf{x}_{\lambda,1} \\
    \vdots\\
    \mathbf{x}_{\lambda,t}
\end{smallmatrix}\right]$) or the output of the Transformer encoder ($\mathbf{X}_\mathrm{enc}$). The attention weights $\mathbf{A}_{\lambda,h}$ are computed as:
\begin{align*}
    \mathbf{A}_{\lambda,h} &= \mathrm{softmax}\left(\frac{\mathbf{Q}_{\lambda,h} \mathbf{K}_{\lambda,h}^\top}{\sqrt{d/H}}\right) \\
    \mathbf{Q}_{\lambda,h} &= \mathbf{W}^{(\mathrm{ma,Q})}_{\lambda,h} \mathbf{X}_\lambda + \mathbf{b}^{(\mathrm{ma,Q})}_{\lambda,h} \\
    \mathbf{K}_{\lambda,h} &= \mathbf{W}^{(\mathrm{ma,K})}_{\lambda,h} \mathbf{X} + \mathbf{b}^{(\mathrm{ma,K})}_{\lambda,h}
\end{align*}
Remark that the matrix $\mathbf{A}_{\lambda,h}$ has size $q \times k$, with $q$ the number of rows in $\mathbf{X}_\lambda$ and $k$ the number of rows in the input $\mathbf{X}$. 
Specific cell values $a_{\lambda hij}$ of this attention matrix $\mathbf{A}_{\lambda,h}$, also known as \emph{attention weights}, can be seen as computing a similarity score for the $i$\textsuperscript{th} (linearly transformed) input contextual embedding and the $j$\textsuperscript{th} (linearly transformed) attended vector.

Lastly, around each sub-layer, a \emph{residual connection} and a \emph{layer-norm} are applied:
\begin{align*}
    \mathbf{e}_{\lambda,t} &= \mathbf{g}^{(\mathrm{ln})}_\lambda \odot \frac{\mathbf{\ddot{e}}_{\lambda,t} - m_{\lambda,t} \vec{1}}{s_{\lambda,t}} + \mathbf{b}^{(\mathrm{ln})}_\lambda \\
    \mathbf{\ddot{e}}_{\lambda,t} &= \mathbf{\dot{e}}_{\lambda,t} + \mathbf{x}_{\lambda,t}
\end{align*}
where $m_{\lambda,t}$ is the mean of the components of $\mathbf{\ddot{e}}_{\lambda,t}$, and  $s_{\lambda,t}$ the corresponding standard deviation.

\section{Supplementary results}

\paragraph{Numerical stability.} Acros all experiments, all decomposed embeddings were tested for numerical stability:
We ensure an absolute tolerance of $\mathrm{tol}_a = 10^{-8}$ and a relative tolerance of $\mathrm{tol}_r = 10^{-5}$, or more formally that the following is true:
\begin{equation*}
\resizebox{0.95\linewidth}{!}{$\forall\mathrm{Dcp} ~\forall \mathbf{e} \qquad \left| ~ \mathbf{e} ~ - \sum\limits_{\mathbf{z} \in \mathrm{Dcp}(\mathbf{e})}\mathbf{z} ~ \right| ~ \leq ~ \mathrm{tol}_a + \mathrm{tol}_r \left| \sum\limits_{\mathbf{z} \in \mathrm{Dcp}(\mathbf{e})} \mathbf{z} ~ \right|$}%
\end{equation*}
In practice, doing so requires 64-bit float precision, despite the models having been trained with fp16.

\subsection{Performance at the corpus level}
\label{adx:sup-res:perf-corpus}

\begin{figure}
    \centering
    \subfloat[$\mathrm{Dcp}_\mathrm{sl}$, $\cos$ and chrF++]{
    \begin{tikzpicture}
    \tikzstyle{every node}=[font=\tiny]
    \begin{axis}[
        ybar, ymin=0, ymax=100,
        symbolic x coords={I,S,T,F,C},
        bar width=.1cm,
        xtick=data,
        xticklabels={$\mathbf{i}$,$\mathbf{s}$,$\mathbf{t}$,$\mathbf{f}$,$\mathbf{c}$},
        enlarge x limits=0.125,
        enlarge y limits={upper=0.6},
        legend style={at={(0.5,1.05)},
            anchor=south,legend columns=-1,
            /tikz/every even column/.append style={column sep=0.5cm}
        },
        nodes near coords always on top/.style={
            scatter/position=absolute,
            positive value/.style={
                at={(axis cs:\pgfkeysvalueof{/data point/x},\pgfkeysvalueof{/data point/y})},
            },
            negative value/.style={
                at={(axis cs:\pgfkeysvalueof{/data point/x},0)},
            },
            every node near coord/.append style={
                check values/.code={%
                    \begingroup
                    \pgfkeys{/pgf/fpu}%
                    \pgfmathparse{\pgfplotspointmeta<0}%
                    \global\let\result=\pgfmathresult
                    \endgroup
                    %
                    %
                    \pgfmathfloatcreate{1}{1.0}{0}%
                    \let\ONE=\pgfmathresult
                    \ifx\result\ONE
                        \pgfkeysalso{/pgfplots/negative value}%
                    \else
                        \pgfkeysalso{/pgfplots/positive value}%
                    \fi
                },
                check values,
                anchor=west,
                rotate=90,
                font=\tiny,
                /pgf/number format/fixed,
                /pgf/number format/zerofill,
                /pgf/number format/precision=1,
                xshift=-0.5ex,
                color=black,
            },
        },
        nodes near coords={
            \pgfmathprintnumber[fixed zerofill,precision=1]{\pgfplotspointmeta}
        },
        nodes near coords always on top,
        height=3.5cm,
        width=1.1\columnwidth,
        xtick align=inside,
        label style={font=\tiny},
        cycle list/Greens-6,
        every axis plot/.append style={
            fill,
        },
    ]

 \addplot+ [draw=black] plot[error bars/.cd, y dir=both, y explicit] coordinates {
(I, 53.18793538919543)
(S, 79.42234930481001)
(T, 47.33581991938786)
(F, 57.67181636692271)
(C, 12.97897283980677)
};

 \addplot+ [draw=black] plot[error bars/.cd, y explicit] coordinates {
(I, 31.07519980081105)
(S, 70.14749746926417)
(T, 38.98263862026774)
(F, 57.64417321420138)
(C, 12.06532967974886)
};

 \addplot+ [draw=black] plot[error bars/.cd, y explicit] coordinates {
(I, 23.03156512427272)
(S, 73.31534449275077)
(T, 27.66026020381303)
(F, 62.29581174364811)
(C, 33.16119551877189)
};

 \addplot+ [draw=black] plot[error bars/.cd, y explicit] coordinates {
(I, 7.782651224077281)
(S, 44.69145027756881)
(T, 9.48523052692975)
(F, 61.97312656683379)
(C, 11.31955789177329)
};

 \addplot+ [draw=black] plot[error bars/.cd, y explicit] coordinates {
(I, 21.508535615025252)
(S, 62.03756668864783)
(T, 4.93844555496327)
(F, 72.21365937378579)
(C, 18.93243095994693)
};

 \addplot+ [draw=black] plot[error bars/.cd, y explicit] coordinates {
(I, 18.21616330112903)
(S, 66.87637847701853)
(T, 27.94422552754935)
(F, 67.6413258626798)
(C, 43.34448673801896)
};

\legend{\tt s0,\tt s1,\tt s2,\tt sla,\tt ine,\tt mul}
    \end{axis}
\end{tikzpicture}
    }
    
    \subfloat[$\mathrm{Dcp}_\mathrm{sl}$, $\mathrm{nr}$ and chrF++]{
    \begin{tikzpicture}
    \tikzstyle{every node}=[font=\tiny]
    \begin{axis}[
        ybar, ymin=0, ymax=100,
        symbolic x coords={I,S,T,F,C},
        bar width=.1cm,
        xtick=data,
        xticklabels={$\mathbf{i}$,$\mathbf{s}$,$\mathbf{t}$,$\mathbf{f}$,$\mathbf{c}$},
        enlarge x limits=0.125,
        enlarge y limits={upper=0.6},
        legend style={at={(0.5,1.05)},
            anchor=south,legend columns=-1,
            /tikz/every even column/.append style={column sep=0.5cm}
        },
        nodes near coords always on top/.style={
            scatter/position=absolute,
            positive value/.style={
                at={(axis cs:\pgfkeysvalueof{/data point/x},\pgfkeysvalueof{/data point/y})},
            },
            negative value/.style={
                at={(axis cs:\pgfkeysvalueof{/data point/x},0)},
            },
            every node near coord/.append style={
                check values/.code={%
                    \begingroup
                    \pgfkeys{/pgf/fpu}%
                    \pgfmathparse{\pgfplotspointmeta<0}%
                    \global\let\result=\pgfmathresult
                    \endgroup
                    %
                    %
                    \pgfmathfloatcreate{1}{1.0}{0}%
                    \let\ONE=\pgfmathresult
                    \ifx\result\ONE
                        \pgfkeysalso{/pgfplots/negative value}%
                    \else
                        \pgfkeysalso{/pgfplots/positive value}%
                    \fi
                },
                check values,
                anchor=west,
                rotate=90,
                font=\tiny,
                /pgf/number format/fixed,
                /pgf/number format/zerofill,
                /pgf/number format/precision=1,
                xshift=-0.5ex,
                color=black,
            },
        },
        nodes near coords={
            \pgfmathprintnumber[fixed zerofill,precision=1]{\pgfplotspointmeta}
        },
        nodes near coords always on top,
        height=3.5cm,
        width=1.1\columnwidth,
        xtick align=inside,
        label style={font=\tiny},
        cycle list/Greens-6,
        every axis plot/.append style={
            fill,
        },
    ]

 \addplot+ [draw=black] plot[error bars/.cd, y dir=both, y explicit] coordinates {
(I, 65.95771796946975)
(S, 50.47580154217371)
(T, 79.68333325862902)
(F, 57.58784125859644)
(C, 62.050289727575766)
};

 \addplot+ [draw=black] plot[error bars/.cd, y explicit] coordinates {
(I, 66.59310111866178)
(S, 47.38434377435932)
(T, 80.73285335618516)
(F, 31.70254718327391)
(C, 63.095593961117814)
};

 \addplot+ [draw=black] plot[error bars/.cd, y explicit] coordinates {
(I, 67.19939615182935)
(S, 34.711181973081814)
(T, 75.25407235228714)
(F, 48.45133124245365)
(C, 64.45779883986108)
};

 \addplot+ [draw=black] plot[error bars/.cd, y explicit] coordinates {
(I, 55.8666082059794)
(S, 26.800279085943004)
(T, 73.45721657947671)
(F, 0.10012326675232)
(C, 65.99937492614576)
};

 \addplot+ [draw=black] plot[error bars/.cd, y explicit] coordinates {
(I, 52.14970506511234)
(S, 68.76084547370255)
(T, 79.16467531311791)
(F, 2.10588095138209)
(C, 66.10601130200398)
};

 \addplot+ [draw=black] plot[error bars/.cd, y explicit] coordinates {
(I, 31.775003395446088)
(S, 55.909545733374074)
(T, 46.30310526152517)
(F, 39.750704061840146)
(C, 72.31806357533124)
};

    \end{axis}
\end{tikzpicture}
    }
    
    \subfloat[$\mathrm{Dcp}_\mathrm{tok}$, $\cos$ and chrF++]{
        \resizebox{0.525\linewidth}{!}{
            \begin{tikzpicture}
    \tikzstyle{every node}=[font=\tiny]
    \begin{axis}[
        ybar, ymin=0, ymax=100,
        symbolic x coords={S,T,C},
        bar width=.1cm,
        xtick=data,
        xticklabels={$\mathbf{s}$,$\mathbf{t}$,$\mathbf{c}$},
        enlarge x limits=0.25,
        enlarge y limits={upper=0.6},
        legend style={at={(0.5,1.05)},
            anchor=south,legend columns=-1,
            /tikz/every even column/.append style={column sep=0.5cm}
        },
        nodes near coords always on top/.style={
            scatter/position=absolute,
            positive value/.style={
                at={(axis cs:\pgfkeysvalueof{/data point/x},\pgfkeysvalueof{/data point/y})},
            },
            negative value/.style={
                at={(axis cs:\pgfkeysvalueof{/data point/x},0)},
            },
            every node near coord/.append style={
                check values/.code={%
                    \begingroup
                    \pgfkeys{/pgf/fpu}%
                    \pgfmathparse{\pgfplotspointmeta<0}%
                    \global\let\result=\pgfmathresult
                    \endgroup
                    %
                    %
                    \pgfmathfloatcreate{1}{1.0}{0}%
                    \let\ONE=\pgfmathresult
                    \ifx\result\ONE
                        \pgfkeysalso{/pgfplots/negative value}%
                    \else
                        \pgfkeysalso{/pgfplots/positive value}%
                    \fi
                },
                check values,
                anchor=west,
                rotate=90,
                font=\tiny,
                /pgf/number format/fixed,
                /pgf/number format/zerofill,
                /pgf/number format/precision=1,
                xshift=-0.5ex,
                color=black,
            },
        },
        nodes near coords={
            \pgfmathprintnumber[fixed zerofill,precision=1]{\pgfplotspointmeta}
        },
        nodes near coords always on top,
        height=3.5cm,
        width=0.65\columnwidth,
        xtick align=inside,
        label style={font=\tiny},
        cycle list/Greens-6,
        every axis plot/.append style={
            fill,
        },
    ]

 \addplot+ [draw=black] plot[error bars/.cd, y dir=both, y explicit] coordinates {
(S, 26.60100520916091)
(T, 0.83573996926966)
(C, 29.05463700541467)
};

 \addplot+ [draw=black] plot[error bars/.cd, y explicit] coordinates {
(S, 24.48365580063853)
(T, 14.736267673250639)
(C, 3.57799745906202)
};

 \addplot+ [draw=black] plot[error bars/.cd, y explicit] coordinates {
(S, 40.884140318926846)
(T, 43.24656550908992)
(C, 17.147608022130658)
};

 \addplot+ [draw=black] plot[error bars/.cd, y explicit] coordinates {
(S, 25.77746563904445)
(T, 27.65071987551054)
(C, 39.42410304770782)
};

 \addplot+ [draw=black] plot[error bars/.cd, y explicit] coordinates {
(S, 12.32265532218841)
(T, 50.49244552977379)
(C, 1.77326015139447)
};

 \addplot+ [draw=black] plot[error bars/.cd, y explicit] coordinates {
(S, 19.30195964546984)
(T, 45.435589182987464)
(C, 15.84370182722622)
};

    \end{axis}
\end{tikzpicture}
        }
    }\hspace{-0.5cm}
    \subfloat[$\mathrm{Dcp}_\mathrm{tok}$, $\mathrm{nr}$ and chrF++]{
        \resizebox{0.475\linewidth}{!}{
            \begin{tikzpicture}
    \tikzstyle{every node}=[font=\tiny]
    \begin{axis}[
        ybar, ymin=0, ymax=100,
        symbolic x coords={S,T,C},
        bar width=.1cm,
        xtick=data,
        yticklabels={},
        xticklabels={$\mathbf{s}$,$\mathbf{t}$,$\mathbf{c}$},
        enlarge x limits=0.25,
        enlarge y limits={upper=0.6},
        legend style={at={(0.5,1.05)},
            anchor=south,legend columns=-1,
            /tikz/every even column/.append style={column sep=0.5cm}
        },
        nodes near coords always on top/.style={
            scatter/position=absolute,
            positive value/.style={
                at={(axis cs:\pgfkeysvalueof{/data point/x},\pgfkeysvalueof{/data point/y})},
            },
            negative value/.style={
                at={(axis cs:\pgfkeysvalueof{/data point/x},0)},
            },
            every node near coord/.append style={
                check values/.code={%
                    \begingroup
                    \pgfkeys{/pgf/fpu}%
                    \pgfmathparse{\pgfplotspointmeta<0}%
                    \global\let\result=\pgfmathresult
                    \endgroup
                    %
                    %
                    \pgfmathfloatcreate{1}{1.0}{0}%
                    \let\ONE=\pgfmathresult
                    \ifx\result\ONE
                        \pgfkeysalso{/pgfplots/negative value}%
                    \else
                        \pgfkeysalso{/pgfplots/positive value}%
                    \fi
                },
                check values,
                anchor=west,
                rotate=90,
                font=\tiny,
                /pgf/number format/fixed,
                /pgf/number format/zerofill,
                /pgf/number format/precision=1,
                xshift=-0.5ex,
                color=black,
            },
        },
        nodes near coords={
            \pgfmathprintnumber[fixed zerofill,precision=1]{\pgfplotspointmeta}
        },
        nodes near coords always on top,
        height=3.5cm,
        width=0.65\columnwidth,
        xtick align=inside,
        label style={font=\tiny},
        cycle list/Greens-6,
        every axis plot/.append style={
            fill,
        },
    ]

 \addplot+ [draw=black] plot[error bars/.cd, y dir=both, y explicit] coordinates {
(S, 23.796352335732)
(T, 57.6634396674827)
(C, 57.41958562570756)
};

 \addplot+ [draw=black] plot[error bars/.cd, y explicit] coordinates {
(S, 13.53770180304533)
(T, 51.23799316766673)
(C, 67.07085064862291)
};

 \addplot+ [draw=black] plot[error bars/.cd, y explicit] coordinates {
(S, 52.55853264208199)
(T, 67.3241349046567)
(C, 69.6211873741382)
};

 \addplot+ [draw=black] plot[error bars/.cd, y explicit] coordinates {
(S, 36.649029652624435)
(T, 13.77162161431475)
(C, 19.86090980859158)
};

 \addplot+ [draw=black] plot[error bars/.cd, y explicit] coordinates {
(S, 62.234891017679104)
(T, 64.42374786958108)
(C, 59.322097270856254)
};

 \addplot+ [draw=black] plot[error bars/.cd, y explicit] coordinates {
(S, 27.582535365942)
(T, 38.08740979609611)
(C, 47.31761490328089)
};

    \end{axis}
\end{tikzpicture}
        }
    }
    \caption{Corpus-level correlation magnitudes (Spearman's $|\rho|$, in \%)  between scalar indicators ($\cos$, $\mathrm{nr}$) and chrF++.}
    \label{fig:corpus-perf:extra:chrf++}
\end{figure}
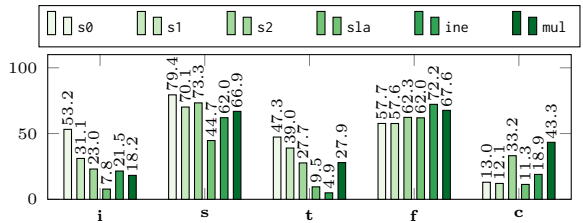
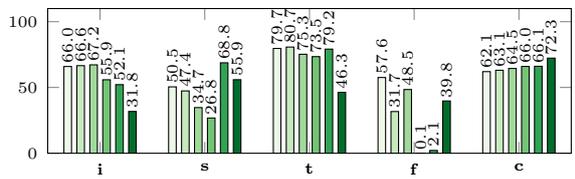
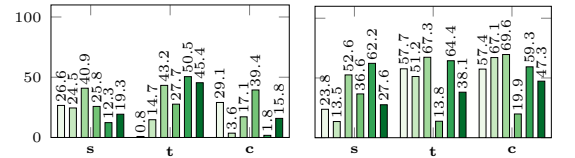

In addition to the BLEU results presented in the main text, we also compute correlation magnitudes using COMET and chrF++ as scoring functions. Corresponding results are presented in \cref{fig:corpus-perf:extra:comet} and \cref{fig:corpus-perf:extra:chrf++}.

\begin{figure}
    \centering
    \subfloat[$\mathrm{Dcp}_\mathrm{sl}$, $\cos$ and COMET]{
    \begin{tikzpicture}
    \tikzstyle{every node}=[font=\tiny]
    \begin{axis}[
        ybar, ymin=0, ymax=100,
        symbolic x coords={I,S,T,F,C},
        bar width=.1cm,
        xtick=data,
        xticklabels={$\mathbf{i}$,$\mathbf{s}$,$\mathbf{t}$,$\mathbf{f}$,$\mathbf{c}$},
        enlarge x limits=0.125,
        enlarge y limits={upper=0.6},
        legend style={at={(0.5,1.05)},
            anchor=south,legend columns=-1,
            /tikz/every even column/.append style={column sep=0.5cm}
        },
        nodes near coords always on top/.style={
            scatter/position=absolute,
            positive value/.style={
                at={(axis cs:\pgfkeysvalueof{/data point/x},\pgfkeysvalueof{/data point/y})},
            },
            negative value/.style={
                at={(axis cs:\pgfkeysvalueof{/data point/x},0)},
            },
            every node near coord/.append style={
                check values/.code={%
                    \begingroup
                    \pgfkeys{/pgf/fpu}%
                    \pgfmathparse{\pgfplotspointmeta<0}%
                    \global\let\result=\pgfmathresult
                    \endgroup
                    %
                    %
                    \pgfmathfloatcreate{1}{1.0}{0}%
                    \let\ONE=\pgfmathresult
                    \ifx\result\ONE
                        \pgfkeysalso{/pgfplots/negative value}%
                    \else
                        \pgfkeysalso{/pgfplots/positive value}%
                    \fi
                },
                check values,
                anchor=west,
                rotate=90,
                font=\tiny,
                /pgf/number format/fixed,
                /pgf/number format/zerofill,
                /pgf/number format/precision=1,
                xshift=-0.5ex,
                color=black,
            },
        },
        nodes near coords={
            \pgfmathprintnumber[fixed zerofill,precision=1]{\pgfplotspointmeta}
        },
        nodes near coords always on top,
        height=4cm,
        width=1.1\columnwidth,
        xtick align=inside,
        label style={font=\tiny},
        cycle list/Reds-6,
        every axis plot/.append style={
            fill,
        },
    ]

 \addplot+ [draw=black] plot[error bars/.cd, y dir=both, y explicit] coordinates {
(I, 55.73862306748022)
(S, 83.59176291748895)
(T, 53.95905096463015)
(F, 64.89879542107349)
(C, 3.9484840876328398)
};

 \addplot+ [draw=black] plot[error bars/.cd, y explicit] coordinates {
(I, 34.22095562319221)
(S, 80.03577983495337)
(T, 52.579099420972916)
(F, 72.61686032245966)
(C, 21.792338805923333)
};

 \addplot+ [draw=black] plot[error bars/.cd, y explicit] coordinates {
(I, 32.9079593503682)
(S, 64.16085028929245)
(T, 43.56551899067643)
(F, 73.26350085077623)
(C, 19.249419017296958)
};

 \addplot+ [draw=black] plot[error bars/.cd, y explicit] coordinates {
(I, 0.50670302716613)
(S, 43.766540207353124)
(T, 17.8047480283829)
(F, 69.46174059672671)
(C, 0.07924637169456)
};

 \addplot+ [draw=black] plot[error bars/.cd, y explicit] coordinates {
(I, 19.39209219287141)
(S, 62.126778033135075)
(T, 5.57389106663816)
(F, 75.70756314103603)
(C, 18.370917058166338)
};

 \addplot+ [draw=black] plot[error bars/.cd, y explicit] coordinates {
(I, 26.463953190006702)
(S, 65.14256396557444)
(T, 22.320435734590973)
(F, 72.37201843082993)
(C, 40.25723313543546)
};

\legend{\tt s0,\tt s1,\tt s2,\tt sla,\tt ine,\tt mul}
    \end{axis}
\end{tikzpicture}
    }
    
    \subfloat[$\mathrm{Dcp}_\mathrm{sl}$, $\mathrm{nr}$ and COMET]{
    \begin{tikzpicture}
    \tikzstyle{every node}=[font=\tiny]
    \begin{axis}[
        ybar, ymin=0, ymax=100,
        symbolic x coords={I,S,T,F,C},
        bar width=.1cm,
        xtick=data,
        xticklabels={$\mathbf{i}$,$\mathbf{s}$,$\mathbf{t}$,$\mathbf{f}$,$\mathbf{c}$},
        enlarge x limits=0.125,
        enlarge y limits={upper=0.6},
        legend style={at={(0.5,1.05)},
            anchor=south,legend columns=-1,
            /tikz/every even column/.append style={column sep=0.5cm}
        },
        nodes near coords always on top/.style={
            scatter/position=absolute,
            positive value/.style={
                at={(axis cs:\pgfkeysvalueof{/data point/x},\pgfkeysvalueof{/data point/y})},
            },
            negative value/.style={
                at={(axis cs:\pgfkeysvalueof{/data point/x},0)},
            },
            every node near coord/.append style={
                check values/.code={%
                    \begingroup
                    \pgfkeys{/pgf/fpu}%
                    \pgfmathparse{\pgfplotspointmeta<0}%
                    \global\let\result=\pgfmathresult
                    \endgroup
                    %
                    %
                    \pgfmathfloatcreate{1}{1.0}{0}%
                    \let\ONE=\pgfmathresult
                    \ifx\result\ONE
                        \pgfkeysalso{/pgfplots/negative value}%
                    \else
                        \pgfkeysalso{/pgfplots/positive value}%
                    \fi
                },
                check values,
                anchor=west,
                rotate=90,
                font=\tiny,
                /pgf/number format/fixed,
                /pgf/number format/zerofill,
                /pgf/number format/precision=1,
                xshift=-0.5ex,
                color=black,
            },
        },
        nodes near coords={
            \pgfmathprintnumber[fixed zerofill,precision=1]{\pgfplotspointmeta}
        },
        nodes near coords always on top,
        height=4cm,
        width=1.1\columnwidth,
        xtick align=inside,
        label style={font=\tiny},
        cycle list/Reds-6,
        every axis plot/.append style={
            fill,
        },
    ]

 \addplot+ [draw=black] plot[error bars/.cd, y dir=both, y explicit] coordinates {
(I, 62.763868021030525)
(S, 49.57831003420065)
(T, 79.54664398647286)
(F, 54.37405452296843)
(C, 69.44772027235108)
};

 \addplot+ [draw=black] plot[error bars/.cd, y explicit] coordinates {
(I, 63.418017218295184)
(S, 30.28529012819395)
(T, 85.93055394551034)
(F, 33.60145947784065)
(C, 76.91093401220925)
};

 \addplot+ [draw=black] plot[error bars/.cd, y explicit] coordinates {
(I, 69.46092970246166)
(S, 43.52822749261037)
(T, 76.37195468029525)
(F, 36.5701125160329)
(C, 73.56825624164661)
};

 \addplot+ [draw=black] plot[error bars/.cd, y explicit] coordinates {
(I, 53.29426277083934)
(S, 25.87815703634445)
(T, 79.7097000969171)
(F, 3.46309192217211)
(C, 66.96024004085722)
};

 \addplot+ [draw=black] plot[error bars/.cd, y explicit] coordinates {
(I, 55.365760855423765)
(S, 67.57770863727211)
(T, 82.84148012309937)
(F, 0.15006024657201)
(C, 67.70681292013496)
};

 \addplot+ [draw=black] plot[error bars/.cd, y explicit] coordinates {
(I, 31.68491665535666)
(S, 54.58375128335252)
(T, 51.65430624460304)
(F, 47.80941925816819)
(C, 70.77032820366476)
};

    \end{axis}
\end{tikzpicture}
    }
    
    \subfloat[$\mathrm{Dcp}_\mathrm{tok}$, $\cos$ and COMET]{
        \resizebox{0.525\linewidth}{!}{
            \begin{tikzpicture}
    \tikzstyle{every node}=[font=\tiny]
    \begin{axis}[
        ybar, ymin=0, ymax=100,
        symbolic x coords={S,T,C},
        bar width=.1cm,
        xtick=data,
        xticklabels={$\mathbf{s}$,$\mathbf{t}$,$\mathbf{c}$},
        enlarge x limits=0.25,
        enlarge y limits={upper=0.6},
        legend style={at={(0.5,1.05)},
            anchor=south,legend columns=-1,
            /tikz/every even column/.append style={column sep=0.5cm}
        },
        nodes near coords always on top/.style={
            scatter/position=absolute,
            positive value/.style={
                at={(axis cs:\pgfkeysvalueof{/data point/x},\pgfkeysvalueof{/data point/y})},
            },
            negative value/.style={
                at={(axis cs:\pgfkeysvalueof{/data point/x},0)},
            },
            every node near coord/.append style={
                check values/.code={%
                    \begingroup
                    \pgfkeys{/pgf/fpu}%
                    \pgfmathparse{\pgfplotspointmeta<0}%
                    \global\let\result=\pgfmathresult
                    \endgroup
                    %
                    %
                    \pgfmathfloatcreate{1}{1.0}{0}%
                    \let\ONE=\pgfmathresult
                    \ifx\result\ONE
                        \pgfkeysalso{/pgfplots/negative value}%
                    \else
                        \pgfkeysalso{/pgfplots/positive value}%
                    \fi
                },
                check values,
                anchor=west,
                rotate=90,
                font=\tiny,
                /pgf/number format/fixed,
                /pgf/number format/zerofill,
                /pgf/number format/precision=1,
                xshift=-0.5ex,
                color=black,
            },
        },
        nodes near coords={
            \pgfmathprintnumber[fixed zerofill,precision=1]{\pgfplotspointmeta}
        },
        nodes near coords always on top,
        height=4cm,
        width=0.65\columnwidth,
        xtick align=inside,
        label style={font=\tiny},
        cycle list/Reds-6,
        every axis plot/.append style={
            fill,
        },
    ]

 \addplot+ [draw=black] plot[error bars/.cd, y dir=both, y explicit] coordinates {
(S, 19.57703587031043)
(T, 9.03383675818665)
(C, 16.25054562429356)
};

 \addplot+ [draw=black] plot[error bars/.cd, y explicit] coordinates {
(S, 17.83147378669298)
(T, 8.31665935484475)
(C, 0.8085795747768301)
};

 \addplot+ [draw=black] plot[error bars/.cd, y explicit] coordinates {
(S, 50.857859152844085)
(T, 51.64912069753368)
(C, 8.12367104424849)
};

 \addplot+ [draw=black] plot[error bars/.cd, y explicit] coordinates {
(S, 23.429942085274792)
(T, 30.72784853624218)
(C, 20.7225392918652)
};

 \addplot+ [draw=black] plot[error bars/.cd, y explicit] coordinates {
(S, 15.771859557548579)
(T, 53.83014304342254)
(C, 0.17979423752202)
};

 \addplot+ [draw=black] plot[error bars/.cd, y explicit] coordinates {
(S, 13.690731131345272)
(T, 40.30218013302726)
(C, 9.05770111581924)
};

    \end{axis}
\end{tikzpicture}
        }
    }\hspace{-0.5cm}
    \subfloat[$\mathrm{Dcp}_\mathrm{tok}$, $\mathrm{nr}$ and COMET]{
        \resizebox{0.475\linewidth}{!}{
            \begin{tikzpicture}
    \tikzstyle{every node}=[font=\tiny]
    \begin{axis}[
        ybar, ymin=0, ymax=100,
        symbolic x coords={S,T,C},
        bar width=.1cm,
        xtick=data,
        yticklabels={},
        xticklabels={$\mathbf{s}$,$\mathbf{t}$,$\mathbf{c}$},
        enlarge x limits=0.25,
        enlarge y limits={upper=0.6},
        legend style={at={(0.5,1.05)},
            anchor=south,legend columns=-1,
            /tikz/every even column/.append style={column sep=0.5cm}
        },
        nodes near coords always on top/.style={
            scatter/position=absolute,
            positive value/.style={
                at={(axis cs:\pgfkeysvalueof{/data point/x},\pgfkeysvalueof{/data point/y})},
            },
            negative value/.style={
                at={(axis cs:\pgfkeysvalueof{/data point/x},0)},
            },
            every node near coord/.append style={
                check values/.code={%
                    \begingroup
                    \pgfkeys{/pgf/fpu}%
                    \pgfmathparse{\pgfplotspointmeta<0}%
                    \global\let\result=\pgfmathresult
                    \endgroup
                    %
                    %
                    \pgfmathfloatcreate{1}{1.0}{0}%
                    \let\ONE=\pgfmathresult
                    \ifx\result\ONE
                        \pgfkeysalso{/pgfplots/negative value}%
                    \else
                        \pgfkeysalso{/pgfplots/positive value}%
                    \fi
                },
                check values,
                anchor=west,
                rotate=90,
                font=\tiny,
                /pgf/number format/fixed,
                /pgf/number format/zerofill,
                /pgf/number format/precision=1,
                xshift=-0.5ex,
                color=black,
            },
        },
        nodes near coords={
            \pgfmathprintnumber[fixed zerofill,precision=1]{\pgfplotspointmeta}
        },
        nodes near coords always on top,
        height=4cm,
        width=0.65\columnwidth,
        xtick align=inside,
        label style={font=\tiny},
        cycle list/Reds-6,
        every axis plot/.append style={
            fill,
        },
    ]

 \addplot+ [draw=black] plot[error bars/.cd, y dir=both, y explicit] coordinates {
(S, 20.376256221271568)
(T, 53.873053223759996)
(C, 53.97380476223348)
};

 \addplot+ [draw=black] plot[error bars/.cd, y explicit] coordinates {
(S, 1.3132749608646899)
(T, 60.72502654209837)
(C, 74.29338782495958)
};

 \addplot+ [draw=black] plot[error bars/.cd, y explicit] coordinates {
(S, 61.62655247277185)
(T, 80.58900794219106)
(C, 83.8468565532571)
};

 \addplot+ [draw=black] plot[error bars/.cd, y explicit] coordinates {
(S, 43.56002019856443)
(T, 2.8096416576558902)
(C, 3.4496067832620603)
};

 \addplot+ [draw=black] plot[error bars/.cd, y explicit] coordinates {
(S, 55.98318028626777)
(T, 59.82759368602262)
(C, 58.338522542547565)
};

 \addplot+ [draw=black] plot[error bars/.cd, y explicit] coordinates {
(S, 30.371969098864348)
(T, 41.80572182879299)
(C, 51.15791882616458)
};

    \end{axis}
\end{tikzpicture}
        }
    }
    \caption{Corpus-level correlation magnitudes (Spearman's $|\rho|$, in \%) between scalar indicators ($\cos$, $\mathrm{nr}$) and COMET.}
    \label{fig:corpus-perf:extra:comet}
\end{figure}
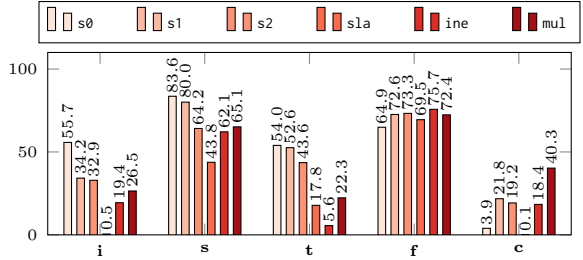
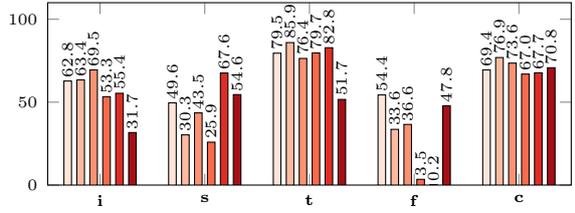
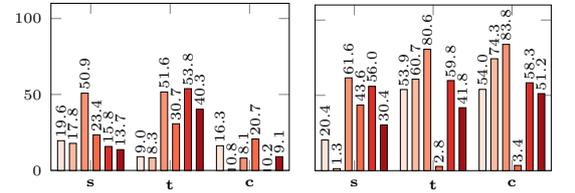

Overall, results are similar to what we observed with BLEU in \cref{fig:corpus-perf}: Setups that yield low or insignificant correlation magnitudes do so across scoring functions.
We nonetheless also attest variation across the different scoring functions, as some specific setups can switch by $\approx 10\%$ depending on the scoring function.

\subsection{Performance at the sentence level}
\label{adx:sup-res:perf-sentence}

In the main body of this article, we measure correlations of sentence-level performance and scalair indicators.
One debatable methodological choice is that we decide to compute signed differences for observations corresponding to the same sentence.

On the one hand, this allows us to factor out some intrinsic variation in scalar indicators that we expect to arise from sheer difference of inputs:
Differences owed to sentence length, idiomaticity, and so on might influence observations---which is why the main results we present do control for input.

On the other hand, one can argue that some inputs will be inherently poorly handled by a model, regardless of its geometry, simply due to training conditions.
Consider for instance a model that would have been solely trained on a bi-text derived from subtitles: Its performances on data derived from parliamentary debates will likely remain low regardless of whether it converges on an efficient set of parameters for its training data.
More succinctly put, one can argue that distributional shifts may impact a sentence-paired approach such as the one we proposed earlier.

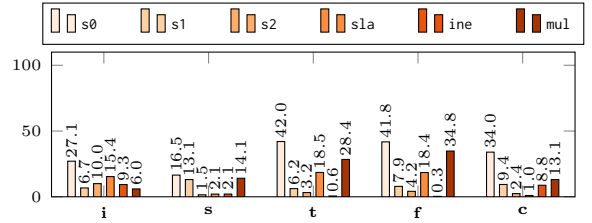
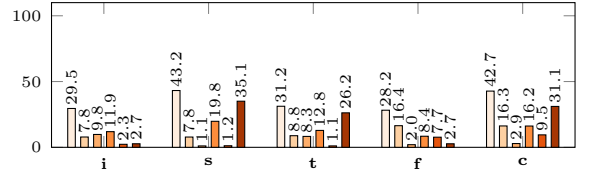
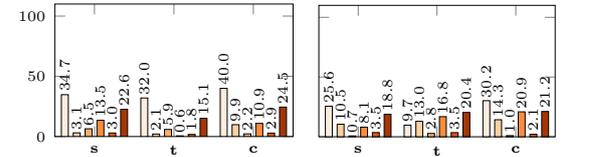
\begin{figure}
    \centering
    \subfloat[$\mathrm{Dcp}_\mathrm{sl}$, $\cos$ and COMET]{
    \begin{tikzpicture}
    \tikzstyle{every node}=[font=\tiny]
    \begin{axis}[
        ybar, ymin=0, ymax=100,
        symbolic x coords={I,S,T,F,C},
        bar width=.1cm,
        xtick=data,
        xticklabels={$\mathbf{i}$,$\mathbf{s}$,$\mathbf{t}$,$\mathbf{f}$,$\mathbf{c}$},
        enlarge x limits=0.125,
        enlarge y limits={upper=0.6},
        legend style={at={(0.5,1.05)},
            anchor=south,legend columns=-1,
            /tikz/every even column/.append style={column sep=0.5cm}
        },
        nodes near coords always on top/.style={
            scatter/position=absolute,
            positive value/.style={
                at={(axis cs:\pgfkeysvalueof{/data point/x},\pgfkeysvalueof{/data point/y})},
            },
            negative value/.style={
                at={(axis cs:\pgfkeysvalueof{/data point/x},0)},
            },
            every node near coord/.append style={
                check values/.code={%
                    \begingroup
                    \pgfkeys{/pgf/fpu}%
                    \pgfmathparse{\pgfplotspointmeta<0}%
                    \global\let\result=\pgfmathresult
                    \endgroup
                    %
                    %
                    \pgfmathfloatcreate{1}{1.0}{0}%
                    \let\ONE=\pgfmathresult
                    \ifx\result\ONE
                        \pgfkeysalso{/pgfplots/negative value}%
                    \else
                        \pgfkeysalso{/pgfplots/positive value}%
                    \fi
                },
                check values,
                anchor=west,
                rotate=90,
                font=\tiny,
                /pgf/number format/fixed,
                /pgf/number format/zerofill,
                /pgf/number format/precision=1,
                xshift=-0.5ex,
                color=black,
            },
        },
        nodes near coords={
            \pgfmathprintnumber[fixed zerofill,precision=1]{\pgfplotspointmeta}
        },
        nodes near coords always on top,
        height=3.5cm,
        width=1.1\columnwidth,
        xtick align=inside,
        label style={font=\tiny},
        cycle list/Oranges-6,
        every axis plot/.append style={
            fill,
        },
    ]

 \addplot+ [draw=black] plot[error bars/.cd, y dir=both, y explicit] coordinates {
(I, 27.07146769264815)
(S, 16.491907005009978)
(T, 41.993086891542816)
(F, 41.7535413644119)
(C, 33.95534990772498)
};

 \addplot+ [draw=black] plot[error bars/.cd, y explicit] coordinates {
(I, 6.72336877809702)
(S, 13.10470279111029)
(T, 6.21514830243121)
(F, 7.91351736463594)
(C, 9.39004860587389)
};

 \addplot+ [draw=black] plot[error bars/.cd, y explicit] coordinates {
(I, 10.00123522123255)
(S, 1.50586501234327)
(T, 3.2097203579867095)
(F, 4.18077084074395)
(C, 2.41467214843302)
};

 \addplot+ [draw=black] plot[error bars/.cd, y explicit] coordinates {
(I, 15.36151939072794)
(S, 2.05301020531329)
(T, 18.52898917655922)
(F, 18.44565802005804)
(C, 1.00482816807364)
};

 \addplot+ [draw=black] plot[error bars/.cd, y explicit] coordinates {
(I, 9.31717349573811)
(S, 2.08380089242007)
(T, 0.61610355660369)
(F, 0.33509677059744)
(C, 8.78014954711006)
};

 \addplot+ [draw=black] plot[error bars/.cd, y explicit] coordinates {
(I, 5.97729254126887)
(S, 14.088445622755568)
(T, 28.40798011633316)
(F, 34.78192375740465)
(C, 13.136823252972722)
};

\legend{\tt s0,\tt s1,\tt s2,\tt sla,\tt ine,\tt mul}
    \end{axis}
\end{tikzpicture}
    }
    
    \subfloat[$\mathrm{Dcp}_\mathrm{sl}$, $\mathrm{nr}$ and COMET]{
    \begin{tikzpicture}
    \tikzstyle{every node}=[font=\tiny]
    \begin{axis}[
        ybar, ymin=0, ymax=100,
        symbolic x coords={I,S,T,F,C},
        bar width=.1cm,
        xtick=data,
        xticklabels={$\mathbf{i}$,$\mathbf{s}$,$\mathbf{t}$,$\mathbf{f}$,$\mathbf{c}$},
        enlarge x limits=0.125,
        enlarge y limits={upper=0.6},
        legend style={at={(0.5,1.05)},
            anchor=south,legend columns=-1,
            /tikz/every even column/.append style={column sep=0.5cm}
        },
        nodes near coords always on top/.style={
            scatter/position=absolute,
            positive value/.style={
                at={(axis cs:\pgfkeysvalueof{/data point/x},\pgfkeysvalueof{/data point/y})},
            },
            negative value/.style={
                at={(axis cs:\pgfkeysvalueof{/data point/x},0)},
            },
            every node near coord/.append style={
                check values/.code={%
                    \begingroup
                    \pgfkeys{/pgf/fpu}%
                    \pgfmathparse{\pgfplotspointmeta<0}%
                    \global\let\result=\pgfmathresult
                    \endgroup
                    %
                    %
                    \pgfmathfloatcreate{1}{1.0}{0}%
                    \let\ONE=\pgfmathresult
                    \ifx\result\ONE
                        \pgfkeysalso{/pgfplots/negative value}%
                    \else
                        \pgfkeysalso{/pgfplots/positive value}%
                    \fi
                },
                check values,
                anchor=west,
                rotate=90,
                font=\tiny,
                /pgf/number format/fixed,
                /pgf/number format/zerofill,
                /pgf/number format/precision=1,
                xshift=-0.5ex,
                color=black,
            },
        },
        nodes near coords={
            \pgfmathprintnumber[fixed zerofill,precision=1]{\pgfplotspointmeta}
        },
        nodes near coords always on top,
        height=3.5cm,
        width=1.1\columnwidth,
        xtick align=inside,
        label style={font=\tiny},
        cycle list/Oranges-6,
        every axis plot/.append style={
            fill,
        },
    ]

 \addplot+ [draw=black] plot[error bars/.cd, y dir=both, y explicit] coordinates {
(I, 29.5307303837968)
(S, 43.17905007989321)
(T, 31.21736745824154)
(F, 28.209130410993332)
(C, 42.747890036210165)
};

 \addplot+ [draw=black] plot[error bars/.cd, y explicit] coordinates {
(I, 7.84443747806929)
(S, 7.8148976660751694)
(T, 8.83445689226855)
(F, 16.44320372524804)
(C, 16.25456869353093)
};

 \addplot+ [draw=black] plot[error bars/.cd, y explicit] coordinates {
(I, 9.7765825913814)
(S, 1.0641541914315)
(T, 8.29608581720668)
(F, 2.0029796817519396)
(C, 2.85990244178496)
};

 \addplot+ [draw=black] plot[error bars/.cd, y explicit] coordinates {
(I, 11.90796184284457)
(S, 19.77592957452009)
(T, 12.848593942152869)
(F, 8.42422022775513)
(C, 16.1822171551682)
};

 \addplot+ [draw=black] plot[error bars/.cd, y explicit] coordinates {
(I, 2.28992817377488)
(S, 1.19539429922066)
(T, 1.07783897860879)
(F, 7.69627381507277)
(C, 9.46882766707659)
};

 \addplot+ [draw=black] plot[error bars/.cd, y explicit] coordinates {
(I, 2.70165710715722)
(S, 35.05137594758306)
(T, 26.2109881306824)
(F, 2.69031706550581)
(C, 31.09012705071057)
};

    \end{axis}
\end{tikzpicture}
    }
    
    \subfloat[$\mathrm{Dcp}_\mathrm{tok}$, $\cos$ and COMET]{
        \resizebox{0.525\linewidth}{!}{
            \begin{tikzpicture}
    \tikzstyle{every node}=[font=\tiny]
    \begin{axis}[
        ybar, ymin=0, ymax=100,
        symbolic x coords={S,T,C},
        bar width=.1cm,
        xtick=data,
        xticklabels={$\mathbf{s}$,$\mathbf{t}$,$\mathbf{c}$},
        enlarge x limits=0.25,
        enlarge y limits={upper=0.6},
        legend style={at={(0.5,1.05)},
            anchor=south,legend columns=-1,
            /tikz/every even column/.append style={column sep=0.5cm}
        },
        nodes near coords always on top/.style={
            scatter/position=absolute,
            positive value/.style={
                at={(axis cs:\pgfkeysvalueof{/data point/x},\pgfkeysvalueof{/data point/y})},
            },
            negative value/.style={
                at={(axis cs:\pgfkeysvalueof{/data point/x},0)},
            },
            every node near coord/.append style={
                check values/.code={%
                    \begingroup
                    \pgfkeys{/pgf/fpu}%
                    \pgfmathparse{\pgfplotspointmeta<0}%
                    \global\let\result=\pgfmathresult
                    \endgroup
                    %
                    %
                    \pgfmathfloatcreate{1}{1.0}{0}%
                    \let\ONE=\pgfmathresult
                    \ifx\result\ONE
                        \pgfkeysalso{/pgfplots/negative value}%
                    \else
                        \pgfkeysalso{/pgfplots/positive value}%
                    \fi
                },
                check values,
                anchor=west,
                rotate=90,
                font=\tiny,
                /pgf/number format/fixed,
                /pgf/number format/zerofill,
                /pgf/number format/precision=1,
                xshift=-0.5ex,
                color=black,
            },
        },
        nodes near coords={
            \pgfmathprintnumber[fixed zerofill,precision=1]{\pgfplotspointmeta}
        },
        nodes near coords always on top,
        height=3.5cm,
        width=0.65\columnwidth,
        xtick align=inside,
        label style={font=\tiny},
        cycle list/Oranges-6,
        every axis plot/.append style={
            fill,
        },
    ]

 \addplot+ [draw=black] plot[error bars/.cd, y dir=both, y explicit] coordinates {
(S, 34.74911850903117)
(T, 32.00252253115367)
(C, 39.99739845588055)
};

 \addplot+ [draw=black] plot[error bars/.cd, y explicit] coordinates {
(S, 3.07541815240518)
(T, 2.06989863705994)
(C, 9.85279078229255)
};

 \addplot+ [draw=black] plot[error bars/.cd, y explicit] coordinates {
(S, 6.462993670873971)
(T, 5.89085944227238)
(C, 2.2347092227921204)
};

 \addplot+ [draw=black] plot[error bars/.cd, y explicit] coordinates {
(S, 13.48968279020516)
(T, 0.58365892803075)
(C, 10.9342710245979)
};

 \addplot+ [draw=black] plot[error bars/.cd, y explicit] coordinates {
(S, 2.96070538049243)
(T, 1.7663007729479498)
(C, 2.86353516212385)
};

 \addplot+ [draw=black] plot[error bars/.cd, y explicit] coordinates {
(S, 22.585786654206828)
(T, 15.112895886015021)
(C, 24.465035123623082)
};

    \end{axis}
\end{tikzpicture}
        }
    }\hspace{-0.5cm}
    \subfloat[$\mathrm{Dcp}_\mathrm{tok}$, $\mathrm{nr}$ and COMET]{
        \resizebox{0.475\linewidth}{!}{
            \begin{tikzpicture}
    \tikzstyle{every node}=[font=\tiny]
    \begin{axis}[
        ybar, ymin=0, ymax=100,
        symbolic x coords={S,T,C},
        bar width=.1cm,
        xtick=data,
        yticklabels={},
        xticklabels={$\mathbf{s}$,$\mathbf{t}$,$\mathbf{c}$},
        enlarge x limits=0.25,
        enlarge y limits={upper=0.6},
        legend style={at={(0.5,1.05)},
            anchor=south,legend columns=-1,
            /tikz/every even column/.append style={column sep=0.5cm}
        },
        nodes near coords always on top/.style={
            scatter/position=absolute,
            positive value/.style={
                at={(axis cs:\pgfkeysvalueof{/data point/x},\pgfkeysvalueof{/data point/y})},
            },
            negative value/.style={
                at={(axis cs:\pgfkeysvalueof{/data point/x},0)},
            },
            every node near coord/.append style={
                check values/.code={%
                    \begingroup
                    \pgfkeys{/pgf/fpu}%
                    \pgfmathparse{\pgfplotspointmeta<0}%
                    \global\let\result=\pgfmathresult
                    \endgroup
                    %
                    %
                    \pgfmathfloatcreate{1}{1.0}{0}%
                    \let\ONE=\pgfmathresult
                    \ifx\result\ONE
                        \pgfkeysalso{/pgfplots/negative value}%
                    \else
                        \pgfkeysalso{/pgfplots/positive value}%
                    \fi
                },
                check values,
                anchor=west,
                rotate=90,
                font=\tiny,
                /pgf/number format/fixed,
                /pgf/number format/zerofill,
                /pgf/number format/precision=1,
                xshift=-0.5ex,
                color=black,
            },
        },
        nodes near coords={
            \pgfmathprintnumber[fixed zerofill,precision=1]{\pgfplotspointmeta}
        },
        nodes near coords always on top,
        height=3.5cm,
        width=0.65\columnwidth,
        xtick align=inside,
        label style={font=\tiny},
        cycle list/Oranges-6,
        every axis plot/.append style={
            fill,
        },
    ]

 \addplot+ [draw=black] plot[error bars/.cd, y dir=both, y explicit] coordinates {
(S, 25.585900958289148)
(T, 9.7466369814922)
(C, 30.209838605607082)
};

 \addplot+ [draw=black] plot[error bars/.cd, y explicit] coordinates {
(S, 10.49584705351088)
(T, 12.961458494726061)
(C, 14.27408296472196)
};

 \addplot+ [draw=black] plot[error bars/.cd, y explicit] coordinates {
(S, 0.73000336915066)
(T, 2.79915130869407)
(C, 0.96445808401463)
};

 \addplot+ [draw=black] plot[error bars/.cd, y explicit] coordinates {
(S, 8.06546775191848)
(T, 16.841608748519548)
(C, 20.883199434438122)
};

 \addplot+ [draw=black] plot[error bars/.cd, y explicit] coordinates {
(S, 3.53093506033436)
(T, 3.4712457452795)
(C, 2.1126236568938297)
};

 \addplot+ [draw=black] plot[error bars/.cd, y explicit] coordinates {
(S, 18.81974387401765)
(T, 20.36908524668126)
(C, 21.162558213140457)
};

    \end{axis}
\end{tikzpicture}
        }
    }
    \caption{Sentence-level correlation magnitudes (Spearman's $|\rho|$, in \%) between scalar indicators ($\cos$, $\mathrm{nr}$) and COMET, without sentence-level pairing.}
    \label{fig:sent-perf:extra}
\end{figure}

We therefore present in \cref{fig:sent-perf:extra} correlation magnitudes derived on unpaired inputs---i.e., we sample two sentences and two checkpoints at random, and compute the corresponding absolute value of the correlation between signed differences.
One can broadly observe two facts:
First, correlation magnitudes are indeed often higher than what we previously reported in \cref{fig:sentence-perf}---however, do recall that one can argue that more variance is expected as we do not control for input variations.
Second, and perhaps more interestingly, we see that whether high correlation magnitudes emerge or not appears highly specific to a given model: in particular, \texttt{s0} and \texttt{mul} almost systematically yields very high correlation magnitudes, whereas other models tend to produce often insignificant scores.

Overall, this supplementary experiment offers an interesting angle: We find evidence that some models' geometry can reflect sentence-level performance, but this does not generalize across different random initializations under the same training conditions.

\end{document}